\documentclass[10pt,journal,compsoc]{IEEEtran}
\ifCLASSOPTIONcompsoc
  \usepackage[nocompress]{cite}
\else
  \usepackage{cite}
\fi
\ifCLASSINFOpdf
 \usepackage[pdftex]{graphicx}
\else
 \usepackage[dvips]{graphicx}
\fi

\usepackage{makecell}
\usepackage{amsmath}
\usepackage{multirow}
\usepackage{amssymb}
\usepackage{bbding}
\usepackage{algorithmic}

\usepackage{array}

\usepackage{gensymb}
\ifCLASSOPTIONcompsoc
 \usepackage[caption=false,font=footnotesize,labelfont=sf,textfont=sf]{subfig}
 
\else
 \usepackage[caption=false,font=footnotesize]{subfig}
\fi

\begin{document}

\title{A Novel Binocular Eye-Tracking System \\With Stereo Stimuli for 3D Gaze Estimation}

\author{Jinglin~Sun,
        Zhipeng~Wu, 
        Han~Wang, 
        Peiguang~Jing,
        Yu~Liu*
\IEEEcompsocitemizethanks{\IEEEcompsocthanksitem J. Sun, Z. Wu, H. Wang and Y. Liu are with the School of Microelectronics, Tianjin University, Tianjin, CN, 30072.\protect\\
E-mail: sunjinglin@tju.edu.cn (J. Sun), liuyu@tju.edu.cn (Y. Liu, $Corresponding$ $Author$)
\IEEEcompsocthanksitem P. Jing is with the School of Electrical and Information, Tianjin University Tianjin, CN, 30072.}
\thanks{Manuscript received April x, 2005; revised August 26, 2015.}}

\markboth{Journal of \LaTeX\ Class Files,~Vol.~xx, No.~x, xx~xx}%
{Shell \MakeLowercase{\textit{et al.}}: Bare Demo of IEEEtran.cls for Computer Society Journals}

\IEEEtitleabstractindextext{%
\begin{abstract}
Eye-tracking technologies have been widely used in applications like psychological studies and human-computer interactions (HCI). 
However, most current eye trackers focus on 2D point-of-gaze (PoG) estimation and cannot provide accurate gaze depth.
Concerning future applications such as HCI with 3D displays, we propose a novel binocular eye-tracking device with stereo stimuli to provide highly accurate 3D PoG estimation.
In our device, the 3D stereo imaging system can provide users with a friendly and immersive 3D visual experience without wearing any accessories. The eye-capturing system can directly record the users' eye movements under 3D stimuli without disturbance.
A regression-based 3D eye-tracking model is built based on collected eye movement data under stereo stimuli. 
Our model estimates users' 2D gaze with features defined by eye region landmarks and further estimates 3D PoG with a multi-source feature set constructed by comprehensive eye movement features and disparity features from stereo stimuli. 
Two test stereo scenes with different depths of field are designed to verify the model’s effectiveness. 
Experimental results show that the average error for 2D gaze estimation was 0.66\degree and for 3D PoG estimation, the average errors are 1.85~cm/0.15~m over the workspace volume 50~cm $\times$ 30~cm $\times$ 75~cm/2.4~m $\times$ 4.0~m $\times$ 7.9~m separately.
\end{abstract}

}

\maketitle

\IEEEdisplaynontitleabstractindextext

\IEEEpeerreviewmaketitle

\IEEEraisesectionheading{\section{Introduction}\label{sec:introduction}}

 \IEEEPARstart{E}{ye}-tracking refers to technologies that can measure and analyze the movements of eyes when persons are observing the world. As an indicator of human visual attention, eye movements are considered to be important clues for nonverbal behaviour analyses~\cite{6327295}. Presently, there have been plenty of eye-tracking-related applications for smart and healthy lives, including psychology studies~\cite{7829437}, marketing and advertising~\cite{morimoto2005eye}, and human-computer interaction (HCI) devices~\cite{ wang2018human,divekar2018cira}.

Eye-tracking technology is promising to be integrated into HCI devices with three-dimensional displays (3D-HCI), as it can extend the current interaction techniques further with eye interactions. Compared with the traditional interacting methods such as speeches or gestures, the human gaze is considered a faster intuitive method that can be used for the 3D-HCI.
For example, eye-tracking technology allows users with motion impairment to intuitively express what they want a robotic device to do by directly looking at the object of interest in the real world~\cite{li20173}.

However, the current major eye-tracking research still focuses on 2D gaze estimation\cite{5557872,8643434,zhang2017deep,zhang2017mpiigaze,wang2019neuro,cheng2020gaze}, and only a few studies have explored the issue of PoG estimation on 3D interfaces~\cite{7164337}. 
Unlike PoG estimation on 2D planes, the 3D PoG estimation needs a new parameter of gaze depth. 
Depth perception is an important feature of 3D vision, and some research about depth perception has been carried out in~\cite{reichelt2010depth,poyade2009influence}.
However, calculation of the gaze depths always encounters limitations, especially in virtual reality~\cite{grinberg1994geometry,akai2007depth}.

Traditional eye-tracking systems usually require an independent external display device, such as a monitor or TV, to display calibration and visual stimulation content. That will cause the eye-tracking system to be larger and cumbersome to operate. The existing eye-tracking systems developed in the current studies can be divided into head-mounted eye trackers~\cite{7164337,7012105,weier2018predicting,hirzle2019design,elmadjian20183d} and remote eye trackers~\cite{kocejko2009eye,gwon2014gaze}. The head-mounted eye trackers usually require users to wear some sort of helmet or glasses to record their eye movement data when they watch videos. Remote eye trackers don't require users to wear any equipment but restrict the users from free head motion.  Once 3D eye-tracking operations have to be performed in stereoscopic environments, the remote eye trackers are not suitable to be used with active shutter glasses~\cite{wibirama2017evaluating}.


To fully explore the huge potential of 3D eye-tracking technology for the 3D-HCI, it is necessary to develop a new integrated device for such research.  
Therefore, in this paper, we propose a novel binocular eye-tracking device with stereo stimuli for 3D PoG estimation, featuring integrated hardware design and efficient algorithm development.
Further, we analyze the key challenges and the future potential in 3D-HCI based on the collected data by our device.

The main contributions of this paper compared to previous works are as follows:
\begin{itemize}

\item A novel integrated binocular eye-tracking system is proposed to solve the current difficulties in gaze estimation on 3D displays. To the best of our knowledge, we are the first to integrate a 3D stereoscopic display with an eye-tracking mechanism into one compact device. 

\item We propose a regression-based eye-tracking model for the 2D and 3D PoG estimation. With the eye movement images and stereo stimulus images as input, this model makes full use of the feature correlation of these two types of data and estimates the users' 2D gaze and 3D PoG based on a multi-source feature set. 

\item In order to fully train and evaluate the 3D PoG estimation model, we design two stereo scenes with different depths of field for tests in virtual and real stereo scenes, respectively. The experimental results show that our model can locate the users' 3D PoG  with high accuracy.

\end{itemize}

The rest of the paper is organized as follows. Section~\ref{Related Work} presents a review of previous works on eye-tracking systems and gaze estimation algorithms. Section~\ref{System Description} presents the proposed system and Section~\ref{3D PoG} presents the 3D PoG estimation model. Section~\ref{experiments} and Section~\ref{results} discuss the workflow of the experiments and results, and Section~\ref{conclusion} concludes this paper.

\section{Related Work}
\label{Related Work}

\subsection{Video-based Eye-tracking System}

\textbf{Eye-Tracking with 2D Stimulus}
Presently, most of the eye trackers proposed by the researchers are oriented to 2D displays. Such devices are usually equipped with a desktop monitor~\cite{gwon2014gaze,hosp2020remoteeye}or a near-eye display~\cite{kocejko2015eye,li2018etracker}.
Gwon et al.~\cite{gwon2014gaze} proposed an eye-tracking system that can be used by subjects who are wearing glasses. To overcome the noise due to reflections from the glasses, they embedded four illuminators at four corners of a monitor, which can be automatically turned on and off sequentially.  Jung et al.~\cite{jung2016compensation} designed an optimal eye-tracking system which was robust to the natural head movements of users.

Kocejko et al.~\cite{kocejko2015eye} proposed an eye-tracking system with a near-eye display, which combined a low-resolution camera with a pair of Google Glasses.     
Li et al.~\cite{li2018etracker} constructed a complete prototype of a  mobile eye-tracking system, ’Etracker’, built with a near-eye viewing device for human gaze tracking. 

More recently, some remote systems have achieved head-pose invariance by explicitly modelling 3D geometries of cameras, eyes, and scenes. Fuchs et al.~\cite{fuchs2019smartlobby} presented an intelligent environment system named SmartLobby.  Benedikt et al.~\cite{hosp2020remoteeye}  presented and evaluated a system named RemoteEye to track such high-speed eye movements. This system could achieve operating frequencies well beyond 500Hz. However, the resolution of the eye image had to be cropped to 100~$\times$~100 pixel to maintain the processing speed.

\textbf{Eye-Tracking with 3D Stimulus} 
Except for a few monocular systems~\cite{lee20123d}, most eye trackers with 3D stimuli are usually binocular systems~\cite{hennessey2008noncontact,abbott2012ultra,wibirama20143d}.
Lee et al.~\cite{lee20123d} developed a monocular 3D gaze tracker using only one eye camera and one NIR illuminator, which was light and convenient for users.
Hennessey et al.~\cite{hennessey2008noncontact} reported and evaluated the first binocular gaze tracking system for estimating the absolute x-, y-, z- coordinates of gaze targets in the real 3D world. However, the system only allowed eyes and head motions within the field of view of the camera. 
In ~\cite{abbott2012ultra}, Abbott reported a 3D head-mounted eye-tracking system, which used two ultra-low-cost video game console cameras at a resolution of 320x240 pixels.
Wibirama et al.~\cite{wibirama20143d} also developed a head-mounted eye-tracking system that can be used properly with consumer-level active shutter glasses.
In this research, the Nvidia 3D Vision system was used to view stereoscopic content.

These existing systems, especially with 3D stimulus, are suitable for the controlled settings but limit their practicality in different fields. For example, head-mounted devices generally restrict the wearing of glasses, and remote devices are difficult to overcome the noise due to reflections from the 3D glasses. Therefore, in this paper, we propose an eye-tracking device with a naked-eye 3D stereo imaging function, which can solve many of the limitations of the above-mentioned system.
Table.~\ref{sys_com} have listed several reported eye-tracking systems with evaluated gaze accuracy in the existing literature.

\begin{table*}[!t]
\setlength\tabcolsep{6pt}
\renewcommand{\arraystretch}{1.5}
\caption{Comparison of Some Recent Relevant Eye-tracking Systems With Our Designed System}
\label{sys_com}
\centering
\begin{tabular}{ccccccccccc}
\hline
Cite & \makecell[c]{Frequency\\(Hz) }& \makecell[c]{ Resolution\\(Pixel)} & Category& \makecell[c]{Binocular\\ camera}& \makecell[c]{Can \\Wearing Glasses}& Stimulus & \makecell[c]{ Average\\Gaze Error}&Workspace\\
\hline
~\cite{gwon2014gaze} &- & 1600~$\times$~1200  &Head-mounted &NO  & YES&2D & 0.70\degree&-\\

~\cite{hosp2020remoteeye} & 575 &100~$\times$~100  & Remote &NO  & YES  &2D& 0.98\degree&- \\

~\cite{li2018etracker} &60 &640~$\times$~480   &Head-mounted & NO & NO&2D & 0.53\degree &-\\

~\cite{jung2016compensation} &30& - & Remote &NO  & YES  &2D & 0.69\degree &- \\

\hline
~\cite{lee20123d} & 30& 640~$\times$~480&Head-mounted &NO  & NO  &3D&5.41~cm&10~$\times$~10~$\times$~50 $c{m^3}$\\

~\cite{hennessey2008noncontact} & 200& 640~$\times$~280&Remote &YES  & YES &3D &3.90~cm &30~$\times$~23~$\times$~25 $c{m^3}$\\


 ~\cite{abbott2012ultra} & 120& 320~$\times$~240&Head-mounted &YES  & NO  &3D& 5.80~cm &47~$\times$~27~$\times$~108 $c{m^3}$ \\

~\cite{wibirama20143d}  &25 &640~$\times$~480  &Head-mounted &YES &YES &3D& 2.10~cm &50~$\times$~28~$\times$~20 $c{m^3}$\\

\hline

\multirow{2}{*}{\textbf{Our}}& \multirow{2}{*}{60}&\multirow{2}{*}{ 1920~$\times$~1080}&\multirow{2}{*}{-}&\multirow{2}{*}{YES}&\multirow{2}{*}{YES} & 2D& 0.66\degree&- \\
& &  &&& &3D& \makecell[c]{1.85~cm\\15.47~cm}& \makecell[c]{ 50~$\times$~30~$\times$~75 $c{m^3}$\\240~$\times$~400~$\times$~700 $c{m^3}$} \\
\hline
\end{tabular}
\end{table*}

\subsection{Gaze Estimation in 3D Space}

The existing methods compute the 3D PoG either use geometric models to calculate the intersection position of the line of sight or build regression mapping or convolutional neural network (CNN) model  based on eye image features.

The visual rays intersection method is considered to be the most direct technique of the 3D PoG estimation. Studies based on this method usually determine the gaze depth by intersecting the visual axes of the two eyes converged on the target~\cite{chamberlain2007eye,daugherty2010measuring,boev2012parameters,hennessey2008noncontact}. Hennessey et al.~\cite{hennessey2008noncontact} reported a binocular eye movement system and have verified this mechanism in a real environment. However, since this technique doesn't rely on any geometry of the 3D scenes, its performance is very sensitive to the errors of the visual rays and the eye positions. Besides, due to angular errors, two visual rays usually don't intersect most of the time, and this error may propagate and accumulate into more severe errors when the visual target is far away from the eyes.

Eye vergence is the most evaluated cue for the 3D PoG estimation in literature. The eye vergence cues can be either horizontal disparity between the 2D gaze points of the left and right eyes~\cite{rozado2013mouse,xia2007ir,li2015gaze}
or the pupil distance~\cite{hennessey2006single,pichitwong20163}.  Recently, some other research also introduced global regression models or machine learning techniques to the vergence-based 3D gaze estimation. For example, Elmadjian et al.~\cite{elmadjian20183d} predicted the X-Y and Z- coordinates of the 3D PoG using two Gaussian process regression (GPR) models. However, these methods must rely on calibration over multiple depths.

More recently, some studies have also explored other ways to estimate the 3D PoG, which are either based on regression or CNN models. For example, Lee et al.~\cite{lee2017estimating} described a gaze depth estimator implemented using a multi-layer perception (MPL) neural network for head-mounted augmented reality (AR) applications. Essig et al.~\cite{essig2006neural} proposed a translation of 2D gazes on screens to 3D positions in virtual spaces with a parameterized self-organizing map (PSOM) network using 27 calibration points. Weier et al.~\cite{weier2018predicting} predicted the gaze depths by combining vergence measures and multiple depth measures into feature sets to train a support vector regression (SVR)model.

\section{Our 3D Eye-Tracking Device}
\subsection{Structure Overview} 
\label{System Description}

\begin{figure}[!t]
\centering
\includegraphics[width=3.5in]{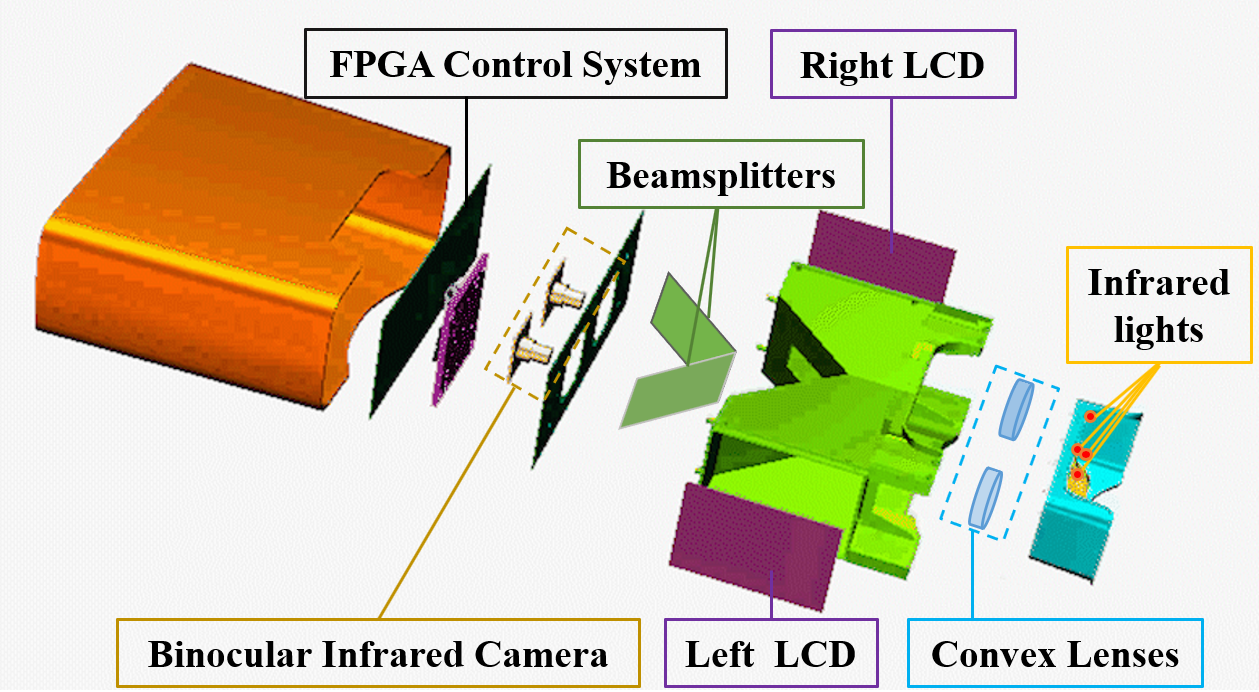}
\caption{Overview of the structure of the designed 3D binocular eye-tracking device. 
}
\label{fig1}
\end{figure}

\begin{figure*}[t!]
\centering
\subfloat[] {\includegraphics[width=3.5in]{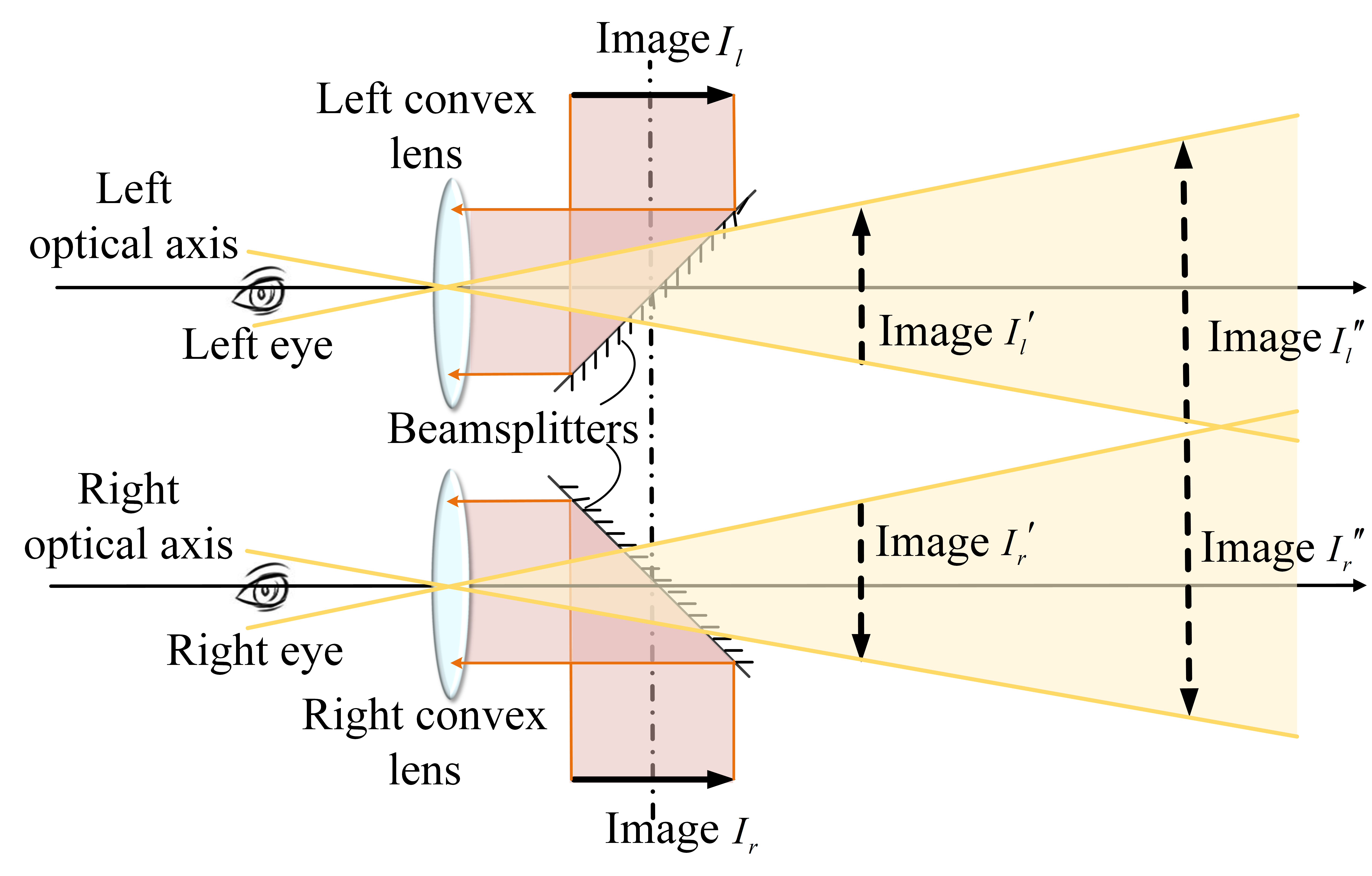}%
\label{fig_2(a)}}
\hfil
\subfloat[]{\includegraphics[width=3.3in]{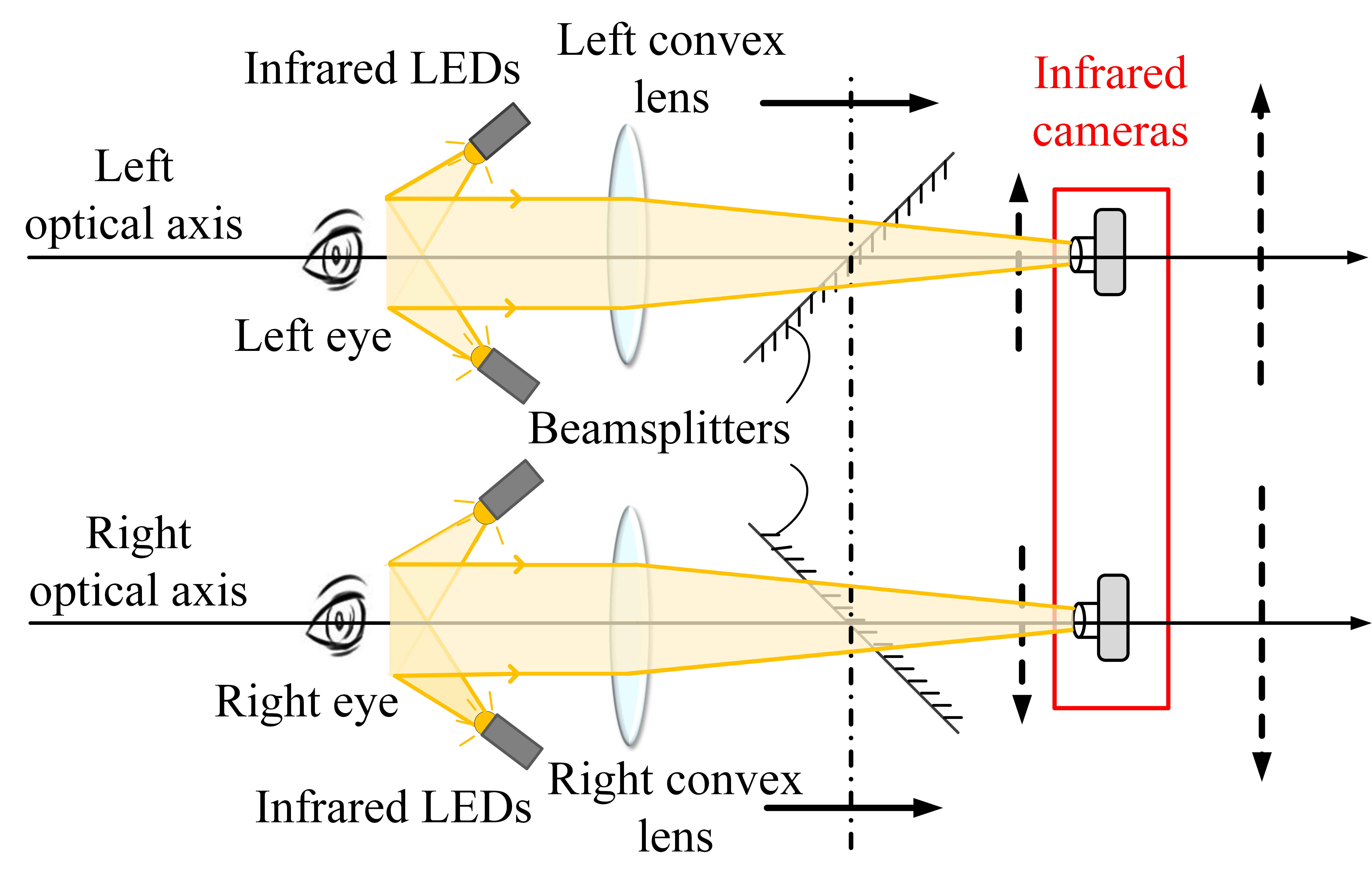}%
\label{fig_2(b)}}
\caption{\textbf {Illustration of the optical paths designed for (a) stereo imaging and (b) eye movement capturing.}}
\label{fig_2}
\end{figure*}

A structural overview of our 3D binocular eye-tracking device is presented in Fig.~\ref{fig1}. The entire device consists of a 3D stereoscopic display system and an eye-tracking system. Both units are tightly integrated with well-designed parameters to allow the stereo imaging and eye-tracking systems to work together without mutual interference.
To ensure the stable operation of the system, we design a flexible FPGA board for system control.  

The device is enclosed into an opaque housing to form a closed field of view to enable users to experience immersive viewing. The 3D stereoscopic display unit is embedded into a carefully designed cavity with a symmetrical structure. This module consists of two liquid crystal display (LCD) screens, two beamsplitters, and two convex lenses. The two LCD screens are erected in parallel and fixed on the left and right sides of the cavity, respectively. Two beamsplitters are fixed inside the cavity at a 45\degree angle to the left and right LCD screens. They can transmit infrared light and reflect visible light. The two compound convex lenses are placed side by side at the front of the cavity to enlarge the image in a certain proportion. As the other main unit, the eye-tracking system includes two infrared cameras and two groups of infrared light sources. Two infrared cameras are installed side by side at the back of the cavity, which can simultaneously collect the binocular eye movement data. Two groups of infrared lights are embedded around the two convex lenses to enhance the infrared exposure of the eye area to allow the cameras to capture eye movement images with higher definition.

The video playback in the 3D stereoscopic display system can be freely switched between 2D and 3D modes. In 3D playback, the left and right LCDs play the left- and right-view images with parallax information. Then, the view images are reflected by the beamsplitters and refracted by the convex lens to enter the users' eyes to form a stereo vision. At the same time, the two infrared cameras located behind the two beamsplitters can collect eye images of the users and track their gaze. 

The main challenge of this design is defining a proper structure that can integrate both systems physically and conform to the relationship of light reflection and transmission, image magnification, and human eye observation. That allows the contents displayed on the left and right LCD screens to overlap to represent a stereo image completely. In addition, our device also features some further advantages. 
i) The bespoke integrated hardware greatly simplifies the complex setup and operating procedures in most eye tracking systems. Users only need to turn on the device with one button when they start using it, and the device can be adjusted in a 360-degree direction  freedmly to meet the needs of different users.
ii) The hardware design avoids the problems of head pose estimation and eye region segmentation, which can effectively save the running time cost and improve the accuracy of the 3D PoG estimation.
iii) The rigorous bespoke software design allows the eye movement data of users' left and right eyes to be collected and stored in our cloud database in a highly synchronized manner.
iv) Unlike some eye trackers restricted in limited stereo scenes, our eye-tracking system allows the change of stereo stimulating scenes when required.

\subsection{Optical Path of Stereo Imaging} 
 In this subsection, we will give detailed analyses on the design of the device's optical paths for the function of stereo imaging.

 The design of optics for stereo imaging has been shown in Fig.~\ref{fig_2(a)}. 
 In 3D playback mode, the control system splits a stereo video clip into two streams of left and right views and displays them on the left and right LCD screens.
 Since both optical paths are symmetrical, we will take the left-view path as an example for frame-based analyses.
 Initially, the source image for the left-view ${I_l}$ displayed on the left LCD can be reflected by the beamsplitter to form a virtual image ${I_l}^\prime $.
 Since the light emitted by the LCD screen is in the visible range, the beamsplitter acts as a plane mirror for the image ${I_l}^\prime $ and the sizes of the visible images ${I_l}$ and ${I_l}^\prime $ are equal.
 In addition, as the angle between the beamsplitter and the optical axis is 45\degree, the virtual image ${I_l}^\prime $ is perpendicular to the optical axis.
 Then, due to the refraction of the convex lens, the virtual image ${I_l}^\prime $ is proportionally enlarged into a virtual image ${I_l}^{\prime \prime }$. 
Finally, for each frame when both of the user’s eyes simultaneously capture the left and right virtual images ${I_l}^{\prime \prime }$, ${I_r}^{\prime \prime }$, the parallax information between the left and right views of each frame of the original stereoscopic video enables the users to generate stereo vision.

From the above analysis, when users watch the stereoscopic
video played by our device, the image plane they are looking
at is the left and right virtual images ${I_l}^{\prime \prime }$, ${I_r}^{\prime \prime }$. In this article, we refer to these planes as the gaze plane.
 
\subsection{Optical Path of Eye Movement Capturing} 
The design of optics for eye movement capturing has been shown  in Fig.~\ref{fig_2(b)}. Considering the illumination provided by the LCDs is in visible light and the brightness is difficult to stabilize, we choose to use infrared light to illuminate the eye regions. Since the left and right eye paths are symmetrical, we also take the left-view optical path as an example to illustrate the design. The light emitted by the infrared sources is reflected by the eye and then enters the convex lens to be refracted into the beamsplitter. As the beamsplitter allows transmission of the infrared light, the reflection from the eye can penetrate the glass and reaches the infrared camera without interfering with the stereo imaging path.

Both types of optical paths are well-designed to enable users to get better experiences during the tests. The setting of the convex lens can magnify and zoom the left- and right- views to a certain ratio, provide users with a more comfortable viewing field, and enable them to obtain a better immersive experience while avoiding eye fatigue. Moreover, to capture comprehensive eye images during tracking, we set the cameras at positions opposing the eyes for direct capture. This design manages to avoid destroying the optical paths for 3D stereo imaging with our designed optics. That design also gives advantages over the existing commercial eye trackers with camera sets fitted under the eyes, which requires camera titling when collecting eye movement data. Such tracking devices are likely to give aliased information due to reflection or viewing angle problems when capturing eye movements from oblique directions, especially when users wear glasses.

\section{Proposed Model for 3D PoG Estimation}
\label{3D PoG}

For eye tracking in real scenes, gaze depth can be derived from the sampled depth at the PoG of the user. However, for 3D displays, due to the virtual nature of the 3D scenes, it is impossible to directly model the virtual scene and get the gaze depth from the 2D on-screen gaze points.
Therefore, this paper converts the 3D PoG estimation under the stereo video stimulus into a regression-based method.

\begin{figure*}[!t]
\centering
\includegraphics[width=6.5in]{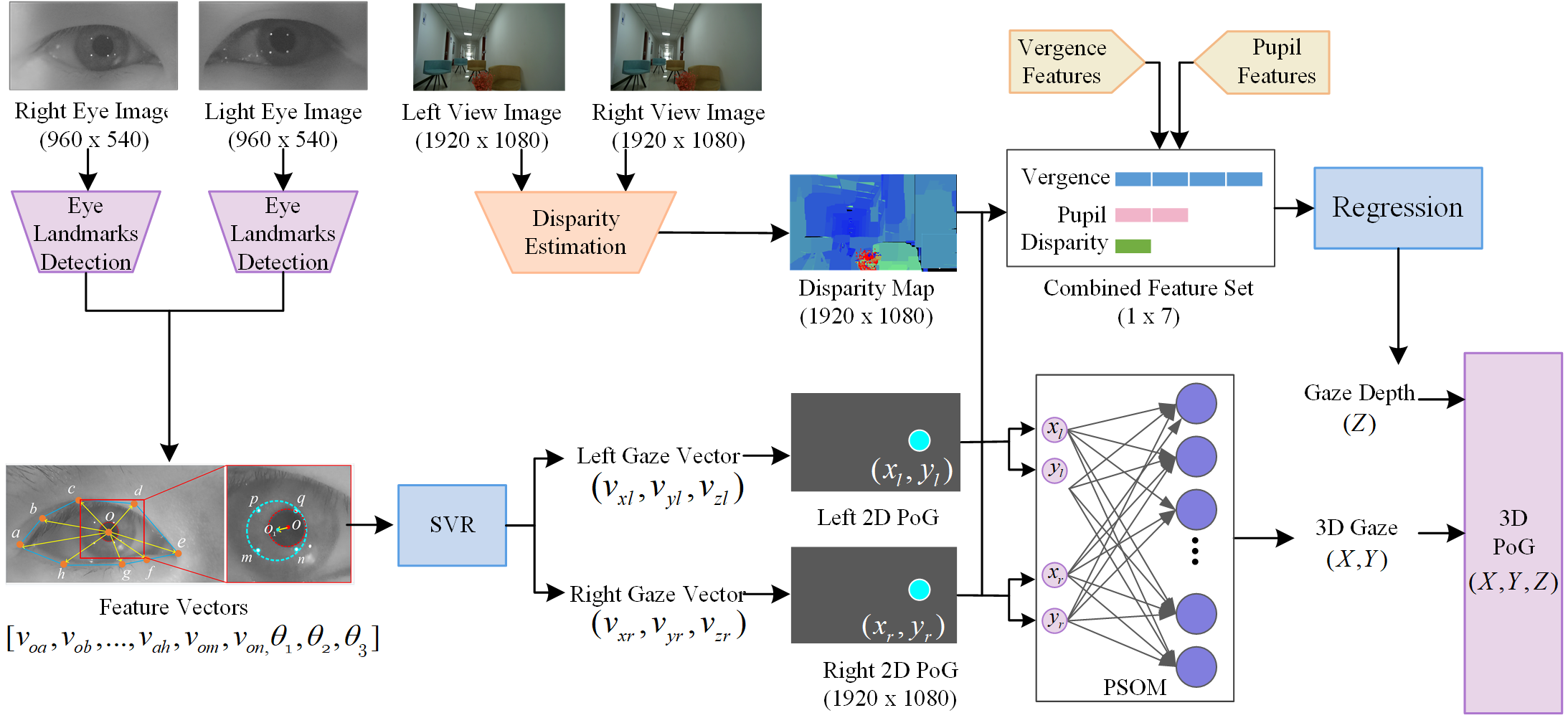}
\caption{The whole workflow of the 3D PoG estimation model.
This model takes eye movement images and stereo scene images as input, and then estimates the coordinates of the user's 3D PoG with constructed feature sets.}
\label{fig4}
\end{figure*}

The whole workflow of the 3D PoG estimation model is shown in Fig.~\ref{fig4}. To construct the feature set required for training of the regression model, we first take the eye movement data and the stereo video stimulation as input to extract the required features. Firstly, some eye region landmarks will be located at the eye images by CNN-based methods, including eyelid, iris, eyeball centres, radius, etc. Then, a feature set based on these eye region landmarks can be built for the gaze direction estimation.
Instead of directly using these landmarks as feature elements, we propose to use the vector values between the centre of the pupil and the other landmarks to build the feature set. Based on this feature set, an SVR model is trained to estimate the users' gaze direction. After that, the 2D gaze points located at the intersection positions of the gaze direction vectors and the gaze planes can be directly computed. To convert the 2D gaze to the 3D space, we train a parameterized self-organizing map (PSOM) network to finish this mapping.
For the gaze depth $Z$ estimation, a
multi-source feature set including eye movement features and
stereo stimuli features is constructed for the regression model's training.  In this step, the disparity value of the stereo view is also introduced as one of the features to improve the relevance of the multi-source feature set. Moreover, to find the optimal solution, five commonly used regression models are trained and evaluated on this feature set to find the optimal solution.  Finally, this whole model can be used for the 3D PoG $(X,Y,Z)$ estimation.

\subsection{Feature Based 2D Gaze Estimation}

As described in the review literature~\cite{hansen2009eye}, feature-based methods for gaze estimation have become the most popular approaches. The primary reasons are that these local eye features such as pupils, eye corners, and eyelids are easy to be extracted from eye images and can be formally related to the gaze. Among them, features based on pupil centre-corneal reflection (PC-CR) vectors~\cite{sigut2010iris} or pupil (iris) center eye corner (PC-EC) vectors~\cite{sesma2012evaluation} have been shown to have a strong correlation with the gaze directions. Inspired by the literature, we propose a more robust feature set to estimate the gaze direction vectors, which makes full use of the relative positions between the pupils and eyelids as well as corneal reflections.

\begin{figure}[!t]
\centering
\includegraphics[width=3.5in]{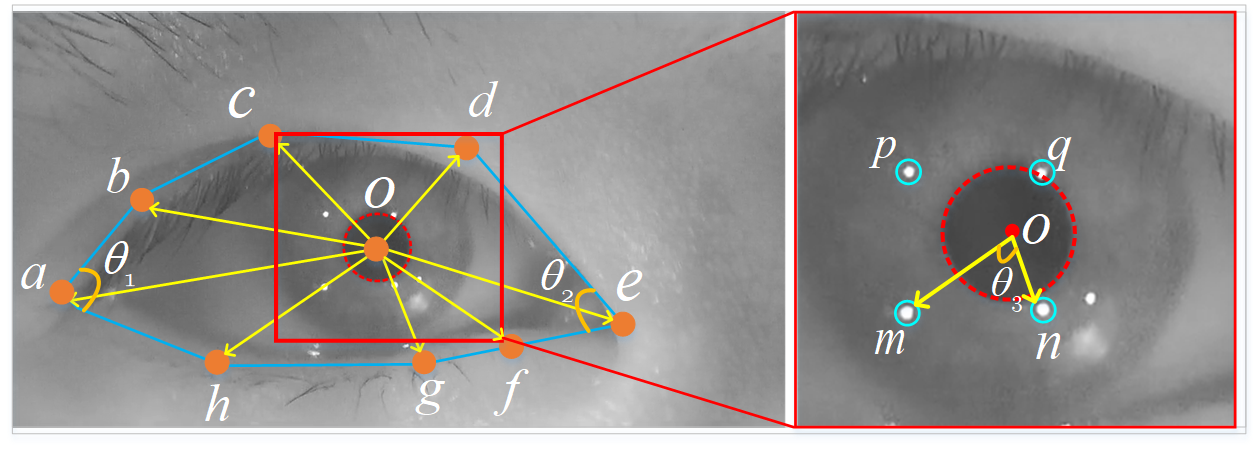}
\caption{ \textbf{Description of the positioning of eye landmarks.}
Symbol $o$ indicates the pupil center landmark, and symbols $(a,b, \dots, h)$ represents eight eyelid landmarks. Symbols $(m,n,p,q)$ marks four corneal reflection points induced by the infrared lights.
}
\label{fig5}
\end{figure}

As shown in Fig.~\ref{fig5}, the symbol $o$ represents the pupil centre landmark and $(a,b, \dots, h)$ represent eyelid landmarks. Symbols $(m,n,p,q)$ represent four corneal reflection points induced by the infrared lights. Symbols ${\theta _1}$, ${\theta _2}$ represent the angles of the inner and outer corners of eye and ${\theta _3}$ represents the angle between two vectors $o \to m$ and $o \to n$. It is worth noting that the two corneal reflection points $p$ and $q$ cannot be guaranteed to be detected at all times. For example, the upper eyelid will block these two points when the user looks down.

Given the above consideration, we choose thirteen features  $[{v_{oa}},{v_{ob}}, \ldots {v_{oh}},{v_{om}},{v_{on}},{\theta _1},{\theta _2},{\theta _3}]$ in total  to form the feature set for a single eye.  In this step, a deep learning method proposed in \cite{park2018learning} is introduced to locate the required eye region landmarks. This method is a robust landmark detector with a stacked-hourglass network which is trained on synthetic eye image datasets. 

With the extracted feature set, an SVR model can be trained for gaze direction vector estimation. Although various models have been proposed to make such predictions, we choose to use the SVR due to its good generalization capability and resilience to over-fitting. Moreover, it has been proved to be very effective in dealing with similar prediction problems~\cite{weier2018predicting,duchowski2011measuring}. Given a labeled training set $\{ ({v_j},{\theta _j}),j = 1,...,N\} $ of $N$  (feature-vector, gaze-vector) pairs, the SVR algorithm learns a  parametric functional mapping from feature vectors to gaze vector estimates of the form:
\begin{equation}
\label{SVR}
\hat \theta (v) = \sum\limits_{j = 1}^N {{\omega _j}\kappa } ({v_j},v) + b,
\end{equation}
where $v$ denotes the extracted feature vector of a test eye image and $b,\,\omega : = {({\omega _1},...,{\omega _N})^T}$ are the parameters of the mapping which are learned from
training data. Besides, $\kappa ( \cdot , \cdot )$ is a chosen positive definite symmetric kernel and we choose the radial-basis function (RBF) as a kernel function in this paper:

\begin{equation}
K({v_j},v) = \exp \left\{ { - {{|{v_j},v{|^2}} \over {{\sigma ^2}}}} \right\}
\end{equation}
where $\sigma $ is a kernel parameter.
Then, the solution can be learned by following the optimization problem: 
\begin{equation}
\label{min}
\mathop {\min }\limits_{b,\omega } \,{1 \over 2}||\omega |{|^2} + C\sum\limits_{j = 1}^N {\max (0,|{\theta _j} - \hat \theta ({v_j})| - \varepsilon )},
\end{equation}
where $C$ is a regularization parameter that controls the trade-off between bias and variance. The parameter $\varepsilon$ controls the trade-off between
sparsity and accuracy with a larger $\varepsilon$ favoring sparser solutions. 

With the extracted feature set, the trained model can directly predict the user's gaze vectors. Then, the 2D gaze of the two eyes in pixels, $({x_l},{y_l})$ and $({x_r},{y_r})$, can be located at the intersections between the gaze vectors ${V_l}$, ${V_r}$ and the gaze planes ${I_l}^{\prime\prime} $, ${I_r}^{\prime\prime} $, respectively (see Fig.~\ref{fig6}).

\subsection{Mapping 2D Gaze to 3D Space}
Since the designed device provides the users with 3D stimulus, we need to map their 2D gaze into 3D space. In this paper, the PSOM network is introduced to finish this mapping.  The PSOM network is one type of artificial neural network widely used in standard 2D eye tracking, and it also has been shown to have the ability to estimate the 3D PoG from a user’s binocular eye-tracking data\cite{essig2006neural}.

Commonly, the operation of the PSOM consists of two components. First, it maps 3D PoG $(X,Y,Z)$ through interpolation onto corresponding 2D gaze based on the calibration data. Second, its recurrent component computes the inverse of this mapping, thereby providing the desired projection from 2D to 3D gaze coordinates\cite{essig2006neural}. However, it is worth mentioning that the calibration process with different depth values is very complicated. For example, the calibration of the $n$-layer depth value needs to contain at least $9 \times n$ calibration points. This is not only a difficult computational challenge but also a challenge for the users to gaze at the calibration points all the time. To make the matter worse, the precision of gaze depth estimation with the PSOM algorithm will decrease rapidly as the depth of the gaze plane increases.
Therefore, in our algorithm, the PSOM algorithm is only used to map of $(X,Y)$ elements in 3D PoG $(X,Y,Z)$. The gaze depth element $Z$ will be estimated using a regression-based method, which will be explained in the next subsection.

Given the above considerations, the PSOM network used in our implementation contains four input neurons, nine inner neurons, and two output neurons, with bidirectional connections between the inner and output layers. 
Since only one depth plane calibration is required in our method, the total calibration points are only 9 positions which are arranged in a $3 \times 3$ grid. 
During the calibration process, both the spatial coordinates and the calibration data for each calibration point are stored into the corresponding inner neuron. In detail, for the ${k^{th}}$ calibration point, the calibration data is provided in the form of a reference vector ${\vec w_k} =({x_{lk}},{y_{lk}},{x_{rk}},{y_{rk}}) \in {R^4}$. Given the $(X,Y)$ coordinates of the current 3D PoG $s$, the PSOM constructs an interpolation function to return the 2D gaze:
\begin{equation}
\label{f(s)}
f(s) = \sum\limits_{k \in A} {H(s,k){{\vec w}_k}},
\end{equation}
where $H$ is an an interpolation basis function.

The inverse function ${f^{ - 1}}$ of $f$, which can approximate $(X,Y)$ coordinates of the 3D PoG from the 2D gaze, can be derived by implementing an
iterative minimization of an error function. This error function is defined as:
\begin{equation}
\label{E(s)}
E(s) = {1 \over 2}{(f(s) - {f_{et}})^2},
\end{equation}
in which $f(s)$ is the 2D gaze data calculated by the PSOM and ${f_{et}}$ is the estimated 2D gaze data. The solution is to be updated in an iterative gradient-descent procedure:
\begin{equation}
\label{E(s)}
s(t + 1) = s(t) - \varepsilon {{\delta E(s)} \over {\delta s}},  \quad with\quad \varepsilon  > 0,
\end{equation}
and the optimal solution will be obtained when $E(s(t))$ falls below a preset threshold value.

\subsection{Gaze Depth Estimation with Multi-Source Features}
To further determine gaze depth element $Z$ of users' 3D PoG, we propose to use a multi-source feature set for regression analysis. As mentioned in~\cite{wang2014online}, relying on a single feature to estimate the gaze depth often causes a large range of errors. For example, depth estimations that rely entirely on eye vergence are usually only accurate for the first 0.5~m to 1~m~\cite{duchowski2011measuring,wang2014online}. Meanwhile, visual rays intersection-based methods for gaze depth prediction have also been proven to be inaccurate for realistic scenes~\cite{mantiuk2011gaze,padmanaban2017optimizing}. Thus, incorporating more depth features from multi source for gaze depth prediction will be beneficial~\cite{weier2018predicting}.

Multiple cues related to gaze depth in human visual perception have been evaluated in literature~\cite{ pfeiffer2008evaluation,reichelt2010depth}, including oculomotor related and visual related cues. In our discussion, however, we only concentrate on the interaction of eye vergence, pupil, and disparity cues because they are more significant and well-matched for our binocular eye-tracking system.

\begin{figure}[!t]
\centering
\includegraphics[width=3in]{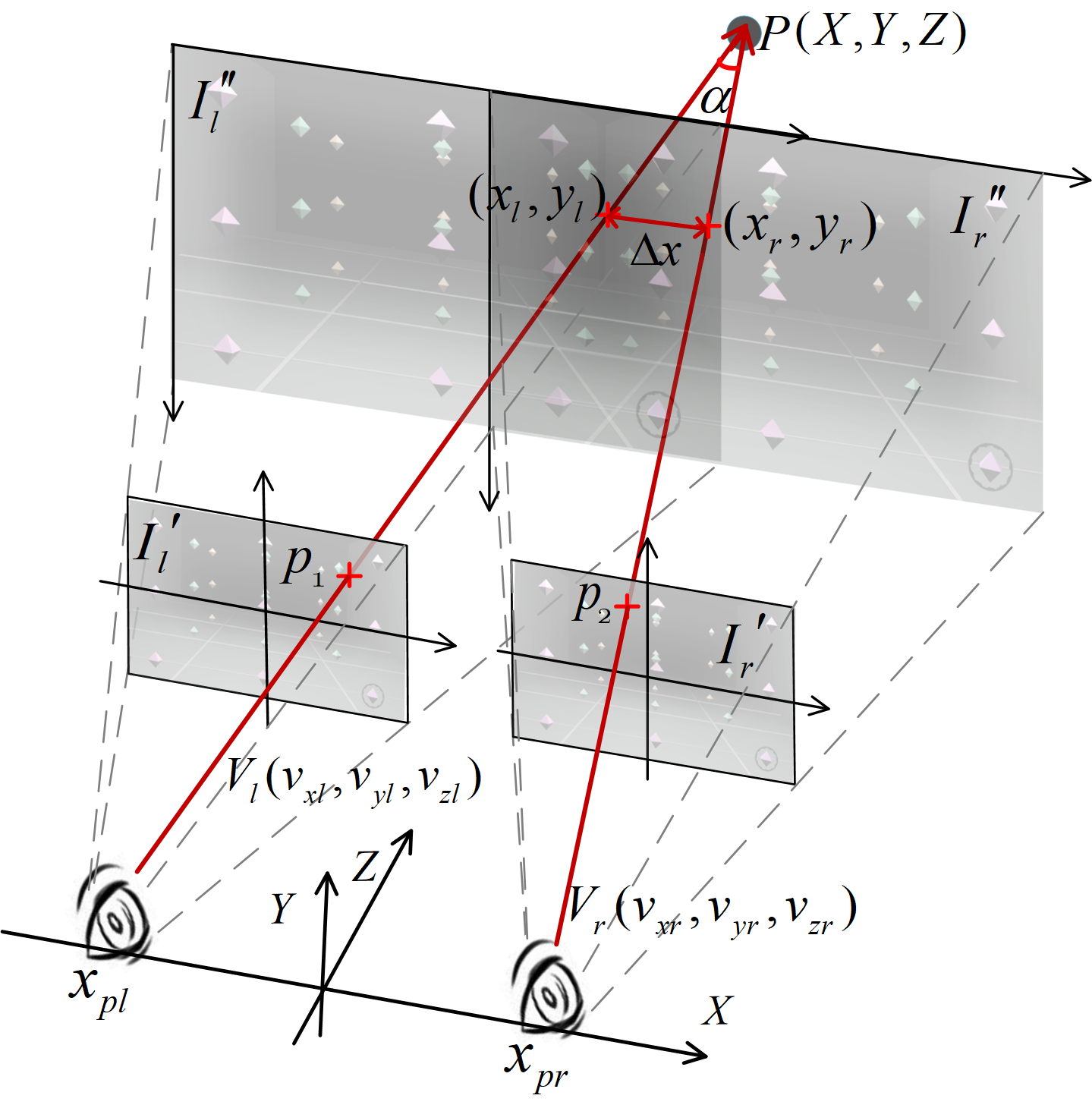}
\caption{\textbf{Description of the multi-source feature set for gaze depth estimation.} Two gaze vectors ${V_l}$ and ${V_r}$ start from the centers of the two pupils, intersect gaze plane at the points ${p_l}$, ${p_2}$, and further intersect at the point $P$ in 3D space forming the vergence angle $\alpha $.}
\label{fig6}
\end{figure}

In Fig.~\ref{fig6}, we give a detailed description of the utilized features. Usually, vergence-related features can be represented as the vergence angle, $\alpha$, and the horizontal vergence, $\Delta x$. The vergence angle is defined as the angle between the two gaze vectors, ${V_l}$ and ${V_r}$. Measuring vergence angles, one may differentiate gaze depths up to a distance of 1.5~m to 3~m depending on the user’s visual faculty~\cite{pfeiffer2008evaluation}. The horizontal disparity $\Delta x$ is described as the horizontal distance of the left and right 2D gaze. The pupil is an additional important factor, and it is extremely sensitive to the amount of received light. Under normal light conditions, its diameter can adjust in a range of 2~$\sim$~6~mm~\cite{campbell1965optical}. As another visual related cue, the disparity of the stereo view can intuitively reflect the depth changes of the 3D stimulus. Theoretically, there is a linear relationship between the disparity value and the depth of field (DoF) value of a stereo view. Besides, as far as we know, we are the first to use disparity features of the stereo image stimuli to estimate the gaze depths.

The multi-source feature set contains a total of 7 columns of features as shown in Table~\ref{Tab2}. Specifically, it consists of three categories: vergence-related features including the x-axis components of the two gaze vectors, $({v_{xl}},{v_{xr}})$, vergence angle, $\alpha $, and horizontal vergence, $\Delta x$; pupil-related features including the interpupillary distance, $|{x_{pr}} - {x_{pl}}|$, and the average pupil radius of two eyes, ${1 \over 2}({R_{l}} + {R_{r}})$; disparity-related features defined as ${D_{(x,y)}}$ that is the value at the 2D gaze point in the disparity map. In order to eliminate the error caused by 2D fixation point estimation, we take $(x,y) = {1 \over 2}({p_1} + {p_2})$, where ${p_1}$ and ${p_2}$ are the intersection points of the gaze vectors and virtual planes ${I_l}^{\prime}$ .

To determine a suitable model, we trained five mainstream regression models for the gaze depth prediction. The five regression models are standard linear regression (SLR), Bayesian ridge regression (BR), elastic network regression (ETCR), SVR, and gradient boosting regression (GBR). In addition, to eliminate errors caused by data fluctuations, these five models were cross-validated for each dataset. We evaluated these models based on the Scikit-learn library, which is a third-party library that encapsulates common machine learning methods in Python. For the sake of fairness, the parameters and kernel functions of each model maintained the initial default values during our evaluation process. The detailed evaluation results are shown in Subsection~\ref{feature analysis}.

This section describes the entire process of building our 3D PoG estimation model without the need to perform multiple depth plane calibration. In the next section, we present the design of several experiments for the evaluation of the model performance.

\begin{table}[!t]
\renewcommand{\arraystretch}{2.0}
\setlength{\tabcolsep}{1.7mm}
\caption{Description of the Multi-Source Feature Set for the Gaze Depth Estimation}
\label{Tab2}
\centering
\begin{tabular}{|c|c|l|}
\hline
Categories & Elements & Description\\
\hline
\multirow{3}{*}{Vergence} &  $({v_{xl}},{v_{xr}})$ & the x-axis components of the gaze vectors\\
\ & $\alpha $ & the angle between the two gaze vectors\\
\ & $\Delta x$ &  the horizontal distance of two gaze points\\
\hline
\multirow{2}{*}{Pupil} & $|{x_{pr}} - {x_{pl}}|$  & the interpupillary distance \\
\ & ${1 \over 2}({R_{l}} + {R_{r}})$ &  the average radius of the two pupils\\
\hline
\ Disparity& ${D_{(x,y)}}$ &  the disparity value at the 2D gaze points\\
\hline
\end{tabular}
\end{table}

\section{DESIGN OF EXPERIMENTS}
\label{experiments}
\subsection{Stereo Stimuli}

\begin{figure}[!t]
\centering
\includegraphics[width=3.5in]{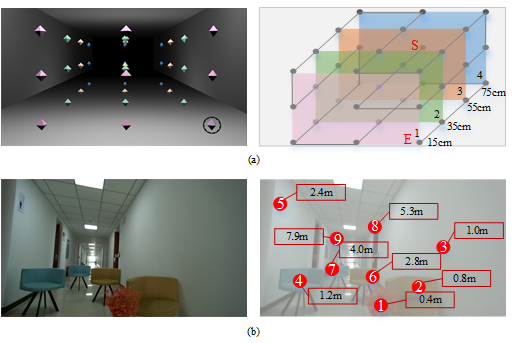}
\caption{\textbf{The two test scenes for 3D gaze model evaluation.} (a) Scene1 is a virtual scene, in which there are a total of $3 \times 3 \times 4$ test points distributed on 4 different depth planes (15~cm, 35~cm, 55~cm, and 75~cm). (b) Scene2 is a realistic indoor scene and there are a total of 9 test points(1$\sim$9) with depths in a range of 0 $\sim$ 7.9~m.}
\label{fig7}
\end{figure}

\begin{figure*}[!t]
\centering
\includegraphics[width=7in]{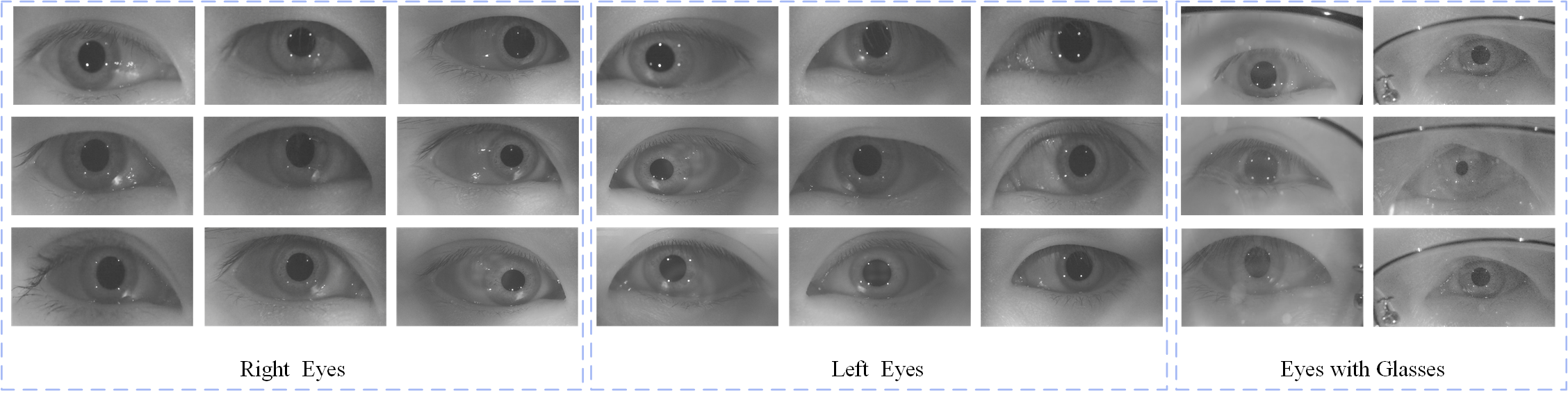}
\caption{Examples of captured eye images, including left eyes, right eyes and eyes with glasses.}
\label{fig9}
\end{figure*}

\begin{figure}[!t]
\centering
\includegraphics[width=3in]{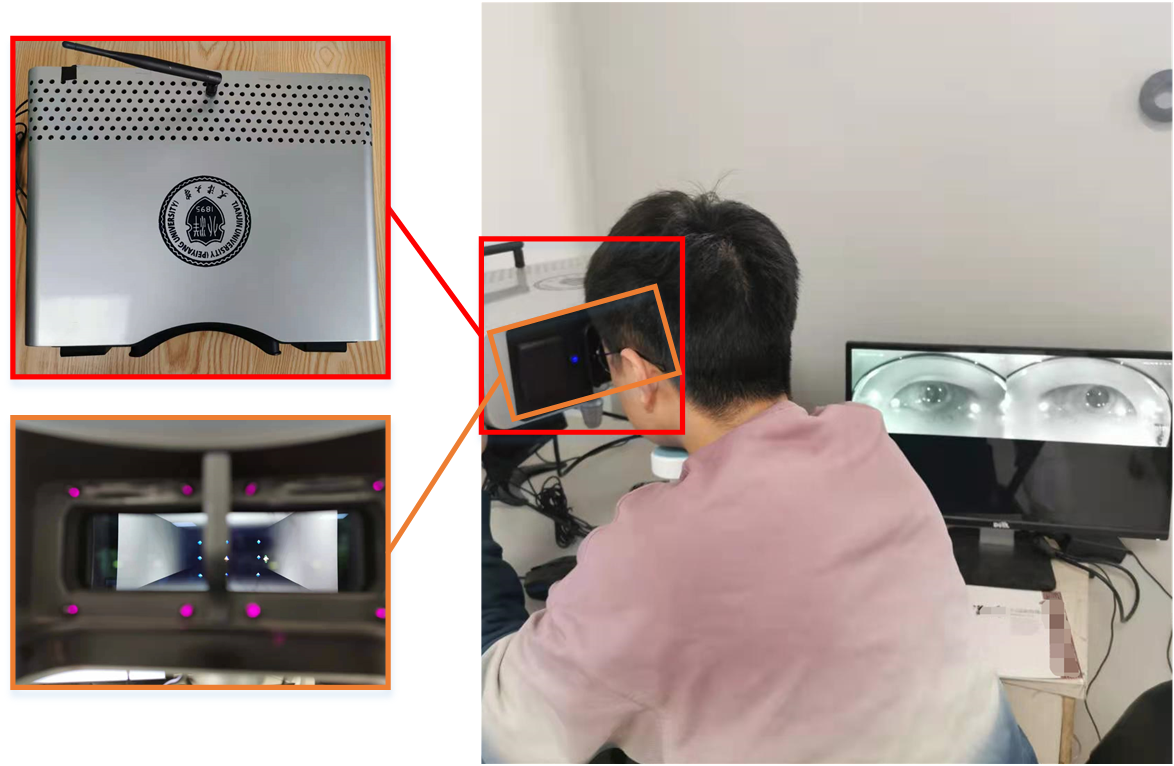}
\caption{\textbf{Experimental setup for data collection.} The complete setup consists of a binocular eye-tracking system and a monitor which can be used to display the user's eye movement data in real time. The head of the user was stabilized using a chin rest.}
\label{fig8}
\end{figure} 

In order to measure the performance of the system for scenes with different volumes, we designed two 3D scenarios with different test volumes.
As shown in Fig.~\ref{fig7}, the first scenario (Scene1) is a virtual scene, which is designed by the 3D Studio MAX software to test model performance in a close-range view. Another scenario (Scene2) is a realistic indoor
scene taken by a binocular stereo camera to test
the model performance in a far-range view.
The test volumes of the two scenarios are $50cm \times 30cm \times 75cm$(Scene1) and $2.4m \times 4.0m \times 7.9m$(Scene2).

Scene1 contains a total of 36 ($3 \times 3 \times 4$) test points, which are distributed on 4 different depth planes, including 15~cm, 35~cm, 55~cm, and 75~cm. 
A ring marker will follow the trajectory during the test to move from point S to point E,
appearing at 9 locations in each plane ($plan{e_1} \sim plan{e_4}$).
Scene2 contains a total of 9 test points with depths in a range of $0 \sim 7.9$~m. During the animation, a red marker will flash randomly at these 9 testing positions with a beep. After the red marker stops flashing, the user needs to continue staring at the prompt positions for two seconds.

\subsection{Procedure of Data Collection}
To verify the effectiveness of the entire system, we have organized 30 people to participate in our study. The participants include 16 males and 14 females, whose ages are between 22 and 35 years old. During the test, the participants did not need to wear any equipment but only to put their chins on a height-adjustable stand (Fig.~\ref{fig8}). 

At the beginning of the experiment, a participant was shown to a 30-second short 3D video clip to experience the form of 3D presentation in advance. Then the device was switched to 2D mode, and the participant was required to complete the calibration process of the 2D gaze. Nine marker positions were arranged in a $3 \times 3$ array, and a red marker appeared randomly at these positions when the calibration started. Once calibration of the 2D gaze was completed, the participant was asked to remain still to watch two 3D scenes (Fig.~\ref{fig7}) and keep their gaze following a prompt target until the videos were over. To reduce users’ fatigue,  we set a 15-second rest after each video playback.

The device can collect and save the users' eye movement images in video mode at 60~Hz with a resolution of 1080P for a single eye. Then, the recorded data were downsampled to reduce data duplication and redundancy.
For each test point, only 8 eye images (four pairs of left and right eyes) of each participant were chosen to build the dataset. We split our dataset to use 25 users' data for training and leave the rest for testing. Some examples of captured eye images are shown in Fig.~\ref{fig9}.

\subsection{Personal 2D Gaze Calibration}

The calibration considers a total of 9 positions. For each marker position, samples are only recorded if eye features and markers are properly recognized by detection algorithms. At this stage, we convert the 2D gaze calibration problem into a non-linear least squares problem. For each sample, $({x_c},{y_c})$ denotes the known coordinate of the calibration point and $(x,y)$
denotes the corresponding gaze point estimated by the SVR model. Letting ${a_0} \sim {a_5}$ and ${b_0} \sim {b_5}$ be unknown coefficients, we utilize the second order polynomial function to map this calibration:
\begin{equation}
\label{RMSE}
{x_c} = {a_0} + {a_1}x + {a_2}y + {a_3}xy + {a_4}{x^2} + {a_5}{y^2},
\end{equation}
\begin{equation}
\label{RMSE}
{y_c} = {b_0} + {b_1}x + {b_2}y + {b_3}xy + {b_4}{x^2} + {b_5}{y^2}.
\end{equation}

When the calibration process is completed, the resulting coefficient vectors are saved and used in real-time during the follow-up eye-tracking process.

\section{EVALUATION AND RESULTS}
\label{results}

\subsection{Evaluation of 2D Gaze Estimation}

\begin{figure}[!t]
\centering
\includegraphics[width=3.5in]{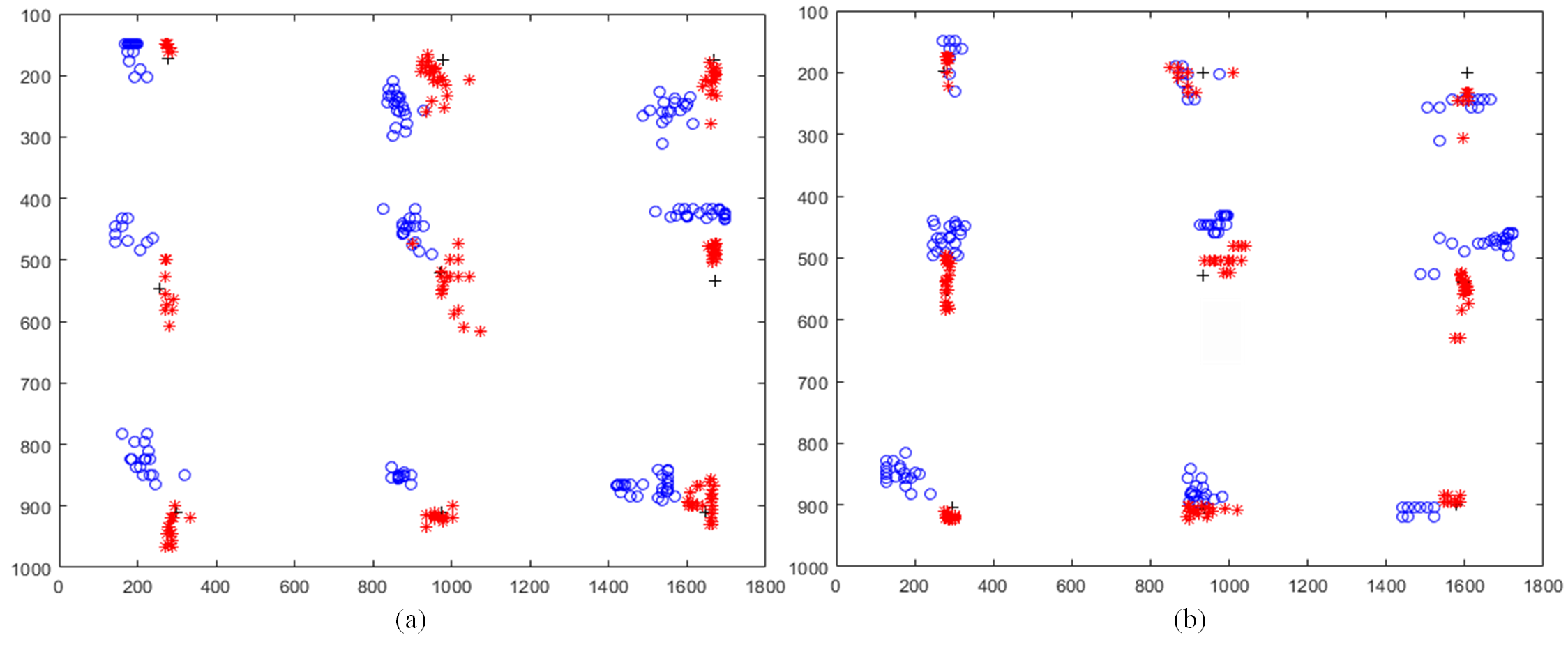}
\caption{\textbf{The calibration results of a user's left (a) and right (b) 2D gaze points.} '$+$' in black represent the true positions of the calibration points, '$o$' in blue represent initially estimated gaze positions, and '$*$' in red represent the calibrated gaze positions.}
\label{fig10}
\end{figure}

We define the 2D gaze estimation errors as the average distances between the estimated 2D gaze points before and after calibration and the true calibration point positions. The estimation results of one user's PoG have been shown in Fig.~\ref{fig10}, where the symbols '$+$' in black represent the true positions of the calibration points, the symbols '$o$' in blue represent estimated the 2D gaze points before calibration, and the symbols '$*$' in red represent the 2D gaze points after calibration. From the calibration result, we can see that the calibration function can reduce the errors of the estimated 2D gaze points to a certain extent. From the statistical analysis of the experimental data, the estimated errors of 2D gaze points before and after calibration are 1.17~cm (1.12\degree) and  0.68~cm (0.66\degree) respectively.

\begin{figure*}[!t]
\centering
\includegraphics[width=6.5in]{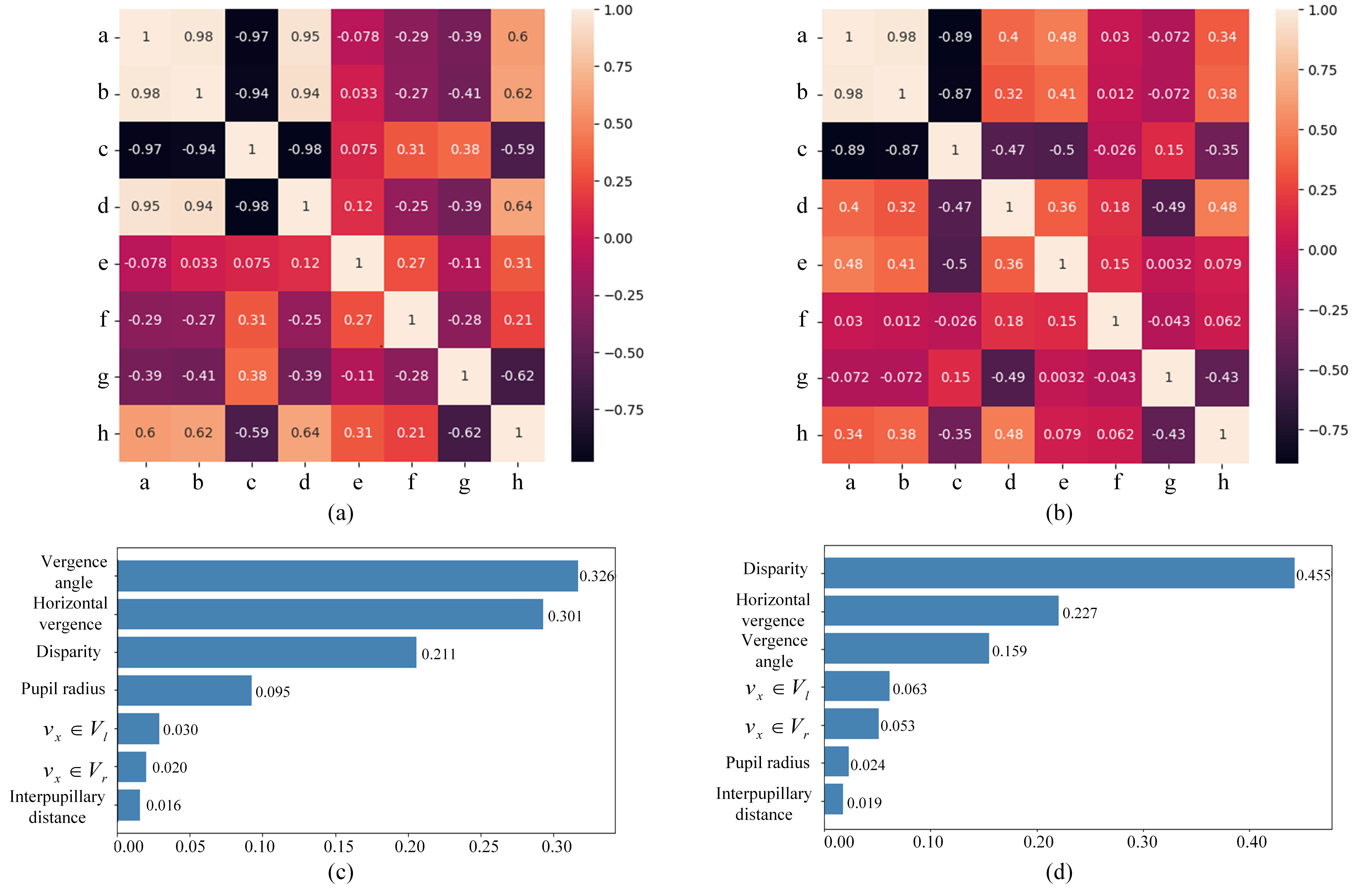}
\caption{\textbf{Linear correlation and importance analysis between each feature vector of the multi-source feature sets.} Figures (a) and (b) show the Pearson correlation coefficient heatmap for the feature sets corresponding to Scene1 and Scene2. Symbols  {a-h} represent the features $ {v_x} \in {V_l}$, ${v_x} \in {V_r} $ vergence angle, horizontal vergence, interpupillary distance, pupil radius, disparity and gaze depth. (c) and (d) show the compute Gini coefficient for the feature sets corresponding to Scene1 and Scene2.}
\label{fig11}
\end{figure*}
 
\begin{figure}[!t]
\centering
\includegraphics[width=3in]{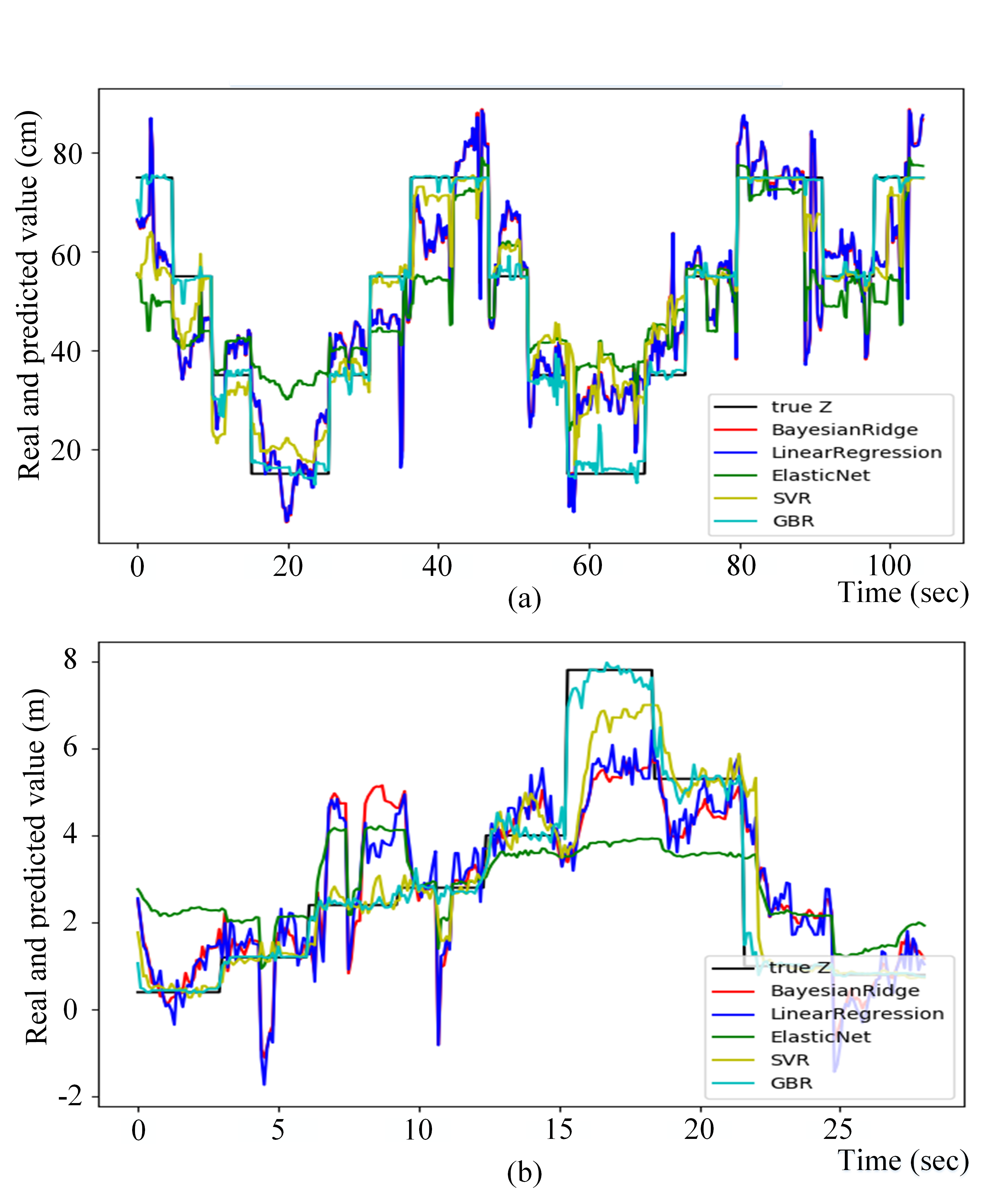}
\caption{\textbf{Comparison of the gaze depth regression results.} (a) and (b) present the comparison in consistency and stability of five models against the ground truths for Scene1 and Scene2, respectively.}
\label{fig12}
\end{figure}

\subsection{Evaluation of $(X,Y)$ Elements}

In this step, we evaluate the accuracy of the estimated $(X,Y)$ elements. The measurement error is defined as the average Euclidean distance from the estimated value to the real gaze-position for individual coordinates $(x-, y-)$. 
Since Scene1 and Scene2 have different test volumes, one is a close-range view and the other is a far-range view, gaze estimation accuracy for these two scenarios will be evaluated separately.

Results for Scene1 are shown in Table~\ref{tab3}. This table shows the average total  error for all subjects, separated for individual planes and coordinates.
In addition, to explore the influence of the distance of the gaze depth on the accuracy of the $(X,Y)$ estimation, we also calculate the average estimation errors for individual planes ($plan{e_1} \sim plan{e_4}$). The statistical analysis of the experimental data shows that the error differences between the four planes are apparent. The average total errors of the furthest plane $plan{e_4}$ in $x-$ and $y-$ directions are 1.43~cm and 1.27~cm, which are 0.87~cm and 0.79~cm higher than those of the closest plane $plan{e_1}$.

Table~\ref{tab33} shows the estimation error for Scene2. Since this scene contains a total of 9 test positions, we have listed the gaze errors for individual users and test points. The statistical analysis of the experimental data shows that the average error for the closest test point (point 1) and the furthest test point (point 9) in $x-$ and $y-$ directions are 3.12~cm and 2.90~cm and 11.21~cm and 12.44~cm. Besides, the average errors of the five users for the whole scene are in $x-$ and $y-$ directions are 6.80~cm and 7.63~cm.  

These results show that the estimation error will decrease as the gaze plane approaches the eyes. The apparent explanation for this effect is that, when the gaze plane is further away from the eyes, the subject's vergence angle changes less during a shift of the 3D gaze points. This is also why the resolution and accuracy of the eye tracker spatial measurements are higher for near than for distant gaze points. 

\begin{table}[!h]
\renewcommand{\arraystretch}{1.5}
\setlength{\tabcolsep}{2.5mm}
\centering
\caption{Average Total Error of the x- and  y-  Coordinates of Scene1 for Individual Planes}
\label{tab3}
 \begin{tabular}{cccccc}
\hline
\multirow{2}{*}{Planes}& \multicolumn{2}{c}{X-}& &  \multicolumn{2}{c}{Y-}\\
\cline{2-3}\cline{5-6}
&\multicolumn{2}{c}{Average total error (cm)}&& \multicolumn{2}{c} {Average total error (cm)}\\
\hline
$ plan{e_1}$ & \multicolumn{2}{c}{$0.56 \pm 0.09$}&&\multicolumn{2}{c}{$0.48 \pm 0.16$} \\
$ plan{e_2}$ & \multicolumn{2}{c}{$0.61 \pm 0.12$}&&\multicolumn{2}{c}{$0.59 \pm 0.11$} \\
$ plan{e_3}$ & \multicolumn{2}{c}{$1.01 \pm 0.14$}&&\multicolumn{2}{c}{$0.96 \pm 0.13$} \\
$ plan{e_4}$ & \multicolumn{2}{c}{$1.43 \pm 0.17$}&&\multicolumn{2}{c}{$1.27 \pm 0.14$} \\
\hline
Average & \multicolumn{2}{c}{$0.90 \pm 0.13$}&&\multicolumn{2}{c}{$0.83 \pm 0.14$} \\
\hline
\end{tabular}
\end{table}

\begin{table*}[!t]
\setlength\tabcolsep{5.5pt}
\renewcommand{\arraystretch}{1.5}
\caption{Average Total Error of the x- and  y-  Coordinates of Scene2 for Individual Testing Point}
\label{tab33}
\centering
 \begin{tabular}{cccccccccccccccccccc}
\hline
& \multicolumn{9}{c}{X- $(cm)$}& &  \multicolumn{9}{c}{Y- $(cm)$}\\
\cline{2-10}\cline{12-20}
 \makecell[c]{Test\\ Points}&1&2&3&4&5&6&7&8&9&&1&2&3&4&5&6&7&8&9\\
\hline
User1&3.12&3.07&5.14&6.07&5.58&8.01&8.78&9.17& 10.22&
&3.21&3.78&5.54&6.01&8.29&8.76&9.17&9.63&11.27  \\
User2&2.07&3.17&5.24&6.13&7.73&8.77&9.01&9.77& 10.09&
&2.21&3.58&5.57&6.78&8.00&9.26&9.78&10.33&12.11  \\
User3&3.10&3.22&4.24&4.37&6.56&8.64&9.23&9.73& 12.36&
&2.73&3.26&4.97&5.99&7.63&9.66&10.18&11.53&13.24  \\
User4&3.62&2.93&4.37&5.75&6.33&8.48&9.64&9.51&10.99&
&3.13&3.66&5.44&6.20&7.92&9.89&10.96&10.43&12.63  \\    
User5&3.71&2.83&4.59&5.92&7.78&9.04&9.95&10.14&11.38&
&3.22&3.42&6.04&6.36&7.31&8.63&11.20&10.76&12.94  \\
\hline
Ave&3.12&3.04&4.72&5.65&6.80&8.73&9.32&9.66&11.21& &2.90&3.54&5.51&6.27&7.96&9.24&10.26&10.54&12.44  \\
\cline{2-10}\cline{12-20}
& \multicolumn{9}{c}{6.80}& &  \multicolumn{9}{c}{7.63}\\
\hline
\end{tabular}
\end{table*}

\subsection{Feature Correlation Analysis}
\label{feature analysis}
Since we construct a multi-source feature set for the gaze depth estimation in this paper, it is necessary to prove the validity of the selected feature data.  For this purpose, the Pearson correlation coefficient and Gini coefficient between each feature and the depth value are calculated. The Pearson correlation coefficient is a linear correlation coefficient used to reflect the degree of linear correlation between two variables. The larger the absolute value, the stronger the correlation. The Gini coefficient is defined as the total decrease in node impurity averaged over all trees of the ensemble, which can be used as a general indicator of feature importance.

\begin{table*}[!t]
\renewcommand{\arraystretch}{1.5}
\setlength{\tabcolsep}{5mm}
\centering
\caption{Evaluation for the Five Regression Models}
\label{Tab4}
 \begin{tabular}{cccccccccc}
\hline
\multirow{2}{*}{Models}
&\multicolumn{4}{c}{Scene1}
&
&\multicolumn{4}{c}{Scene2}
\\
\cline{2-5} \cline{7-10}
& MAE& MSE & ${R^2}$& Error $(cm)$ & & MAE & MSE &  ${R^2}$ & Error $(m)$\\
\hline
LR & 0.7140 & 0.9199 & 0.6773 & 3.3721 && 0.9688 & 1.5719& 0.5929& 0.3410\\
BR &  0.7162 & 0.9199 & 0.6773 & 3.6008&&  0.9891 & 1.6712 & 0.5672& 0.3511  \\
ETCR & 1.2632 & 2.1449 & 0.2476 & 5.1439 && 1.3220 & 2.4360 &0.3692& 0.4141 \\
\textbf{SVR} &\textbf{0.1474} & \textbf{0.1425} & \textbf{0.9100} & \textbf{1.4830}&&\textbf{0.3217}& \textbf{0.5244}& \textbf{0.8642}&\textbf{0.1243} \\
\textbf{GBR} &\textbf{0.0301} & \textbf{0.1050} & \textbf{0.9242} & \textbf{1.4812}&&\textbf{0.1924} & \textbf{0.2451} & \textbf{0.9036}& \textbf{0.1163} \\
\hline
\end{tabular}
\end{table*}

Fig.~\ref{fig11} shows the computed results for the selected feature sets corresponding to Scene1 (Fig.~\ref{fig11}(a)(c)) and Scene2 (Fig.~\ref{fig11}(b)(d)). 
The two Pearson correlation coefficient heatmaps reveal a remarkably high degree of correlation between the gaze depth and the selected features, especially for Scene1, which is the close-range view. Among these seven-dimensional features, the dimension of the vergence-related and disparity-related features have the highest correlation with gaze depth.  Although the correlation values of such features for Scene2 are relatively small compared with those for Scene1, these features still have significant correlations with the gaze depth (especially for the horizontal vergence feature 0.48 and disparity feature -0.43). In addition, by comparing the two heat maps, it is easy to find that the pupil-related features are susceptible to the depth of field value of the test scenes. When the depth of field of the test scenes is too large, the correlations between pupil-related features and gaze depth will tend to be 0. The computed Gini coefficients for the two feature sets are shown in Fig.~\ref{fig11}(c)(d), and the feature importance scores provide a relative ranking. For Scene1, the top three features are vergence angle, horizontal vergence, and disparity, with scores of 0.326, 0.301, and 0.211, respectively. As for Scene2, although the types of the top three features in the feature importance score ranking are the same as that of Scene1, the ranking order has a noticeable change, and the importance score of the disparity feature reaches the top. This result fully proves that, except for the vergence-based feature, the disparity feature is another crucial cue for gaze depth estimation.

\subsection{Evaluation of Gaze Depth Regression}

To find an optimal solution for gaze depth estimation, we evaluated the performance of five regression models based on the two designed scenes. In this part, four metrics including the mean absolute error (MAE), mean squared error (MSE), coefficient of determination ($R^2$), and the prediction error, are used to make this assessment. Additionally, to thoroughly verify the regression performance of these models under different depths of field, we separate the datasets of Scene1 and Scene2 to train and test the models. 

Table~\ref{Tab4} compares the metrics of the trained models both for Scene1 and Scene2, and the data in the error column is defined as the average Euclidean distances between the average predicted gaze depths and the ground truth values. As can be seen from the table, SVR and GBR models outperform most other algorithms. Especially, the GBR model can achieve 0.0301 in MAE, 0.1050 in MSE, 0.9242 in ${R^2}$, and with 1.4812~cm prediction error for Scene1, and 0.1924 in MAE, 0.2451 in MSE, 0.9036 in ${R^2}$, and with 0.1163~m prediction error for Scene2. Moreover, we find that LR and BR present similar performance. The GBR has an advantage over them as it generates a strong learner by integrating weak prediction models, such as Decision Tree (DT), Random Forest (RF), etc., and generating a strong prediction model. The goal of each weak learner is to fit the negative gradient of the loss function of the previously accumulated model so that after adding the weak learner, the accumulated model loss can be reduced in the direction of the negative gradient.

Fig.~\ref{fig12} shows the comparison of the gaze depth regression results. It can be seen that the GBR model performs the best, and its predicted values are highly consistent with the true values. Compared with the GBR model, the regression performance of the SVR model is slightly worse, and the prediction stability is less robust. Moreover, the performances of the remaining three linear models are not ideal, which may be caused by the low linear correlation of the multi-source features.

Moreover, the average error of the 3D PoG estimation in this paper is defined as the Euclidean distance error $\sqrt {(x_e^2 + y_e^2 + z_e^2)}$, where $x_e, y_e, z_e$ are the average absolute errors for the X-, Y-, and Z- coordinates. The experimental results and the comparison with some eye-tracking systems in the existing literature are shown in Table~\ref{sys_com}.

\section{Conclusion}
\label{conclusion}

This paper proposes a novel binocular eye-tracking system that integrates 3D stereo imaging and eye movement capturing to provide 3D PoG estimation with high accuracy. 
Compared with existing solutions, our design can provide more unified and standardized test scenes for 3D gaze estimation with the benefits of cost-effectiveness and user-friendliness.
To locate the user's PoG with high accuracy in the current 3D field of view, we propose a regression-based 3D PoG estimation model.
This model makes full use of the user's eye movement features and the stereo stimuli features to predict the 3D PoG.  
In addition, two test stereo scenes with different depths of field are designed to verify the effectiveness of our model. 
Experimental results show that the average error for 2D gaze estimation was 0.66\degree and for 3D PoG estimation, the average errors are 1.85~cm/0.15~m over the workspace volume 50~cm $\times$ 30~cm $\times$ 75~cm/2.4~m $\times$ 4.0~m $\times$ 7.9~m separately.
To the best of our knowledge, our proposed device is the first compact and practical solution with bespoke integrated hardware and efficient algorithms to provide accurate 3D PoG estimation.

\ifCLASSOPTIONcaptionsoff
  \newpage
\fi


\begin{thebibliography}{53}

\bibitem{6327295}
Gennady Andrienko, Natalia Andrienko, Michael Burch, and Daniel Weiskopf.
\newblock Visual analytics methodology for eye movement studies.
\newblock {\em IEEE Transactions on Visualization and Computer Graphics},
  18(12):2889--2898, 2012.

\bibitem{7829437}
Jason Orlosky, Yuta Itoh, Maud Ranchet, Kiyoshi Kiyokawa, John Morgan, and
  Hannes Devos.
\newblock Emulation of physician tasks in eye-tracked virtual reality for
  remote diagnosis of neurodegenerative disease.
\newblock {\em IEEE Transactions on Visualization and Computer Graphics},
  23(4):1302--1311, 2017.

\bibitem{morimoto2005eye}
Carlos~H Morimoto and Marcio~RM Mimica.
\newblock Eye gaze tracking techniques for interactive applications.
\newblock {\em Computer vision and image understanding}, 98(1):4--24, 2005.

\bibitem{wang2018human}
Kang Wang, Rui Zhao, and Qiang Ji.
\newblock Human computer interaction with head pose, eye gaze and body
  gestures.
\newblock In {\em 2018 13th IEEE International Conference on Automatic Face \&
  Gesture Recognition (FG 2018)}, pages 789--789. IEEE, 2018.

\bibitem{divekar2018cira}
Rahul~R Divekar, Matthew Peveler, Robert Rouhani, Rui Zhao, Jeffrey~O Kephart,
  David Allen, Kang Wang, Qiang Ji, and Hui Su.
\newblock Cira: An architecture for building configurable immersive
  smart-rooms.
\newblock In {\em Proceedings of SAI Intelligent Systems Conference}, pages
  76--95. Springer, 2018.

\bibitem{li20173}
Songpo Li, Xiaoli Zhang, and Jeremy~D Webb.
\newblock 3-d-gaze-based robotic grasping through mimicking human visuomotor
  function for people with motion impairments.
\newblock {\em IEEE Transactions on Biomedical Engineering}, 64(12):2824--2835,
  2017.

\bibitem{5557872}
Hiroki Mori, Erika Sumiya, Tomohiro Mashita, Kiyoshi Kiyokawa, and Haruo
  Takemura.
\newblock A wide-view parallax-free eye-mark recorder with a hyperboloidal
  half-silvered mirror and appearance-based gaze estimation.
\newblock {\em IEEE Transactions on Visualization and Computer Graphics},
  17(7):900--912, 2011.

\bibitem{8643434}
Zhiming Hu, Congyi Zhang, Sheng Li, Guoping Wang, and Dinesh Manocha.
\newblock Sgaze: A data-driven eye-head coordination model for real time gaze
  prediction.
\newblock {\em IEEE Transactions on Visualization and Computer Graphics},
  25(5):2002--2010, 2019.

\bibitem{zhang2017deep}
Mengmi Zhang, Keng Teck~Ma, Joo Hwee~Lim, Qi~Zhao, and Jiashi Feng.
\newblock Deep future gaze: Gaze anticipation on egocentric videos using
  adversarial networks.
\newblock In {\em Proceedings of the IEEE conference on computer vision and
  pattern recognition}, pages 4372--4381, 2017.

\bibitem{zhang2017mpiigaze}
Xucong Zhang, Yusuke Sugano, Mario Fritz, and Andreas Bulling.
\newblock Mpiigaze: Real-world dataset and deep appearance-based gaze
  estimation.
\newblock {\em IEEE transactions on pattern analysis and machine intelligence},
  41(1):162--175, 2017.

\bibitem{wang2019neuro}
Kang Wang, Hui Su, and Qiang Ji.
\newblock Neuro-inspired eye tracking with eye movement dynamics.
\newblock In {\em Proceedings of the IEEE/CVF Conference on Computer Vision and
  Pattern Recognition}, pages 9831--9840, 2019.

\bibitem{cheng2020gaze}
Yihua Cheng, Xucong Zhang, Feng Lu, and Yoichi Sato.
\newblock Gaze estimation by exploring two-eye asymmetry.
\newblock {\em IEEE Transactions on Image Processing}, 29:5259--5272, 2020.


\bibitem{7164337}
Jason Orlosky, Takumi Toyama, Kiyoshi Kiyokawa, and Daniel Sonntag.
\newblock Modular: Eye-controlled vision augmentations for head mounted
  displays.
\newblock {\em IEEE Transactions on Visualization and Computer Graphics},
  21(11):1259--1268, 2015.

\bibitem{reichelt2010depth}
Stephan Reichelt, Ralf H{\"a}ussler, Gerald F{\"u}tterer, and Norbert Leister.
\newblock Depth cues in human visual perception and their realization in 3d
  displays.
\newblock In {\em Three-Dimensional Imaging, Visualization, and Display 2010
  and Display Technologies and Applications for Defense, Security, and Avionics
  IV}, volume 7690, page 76900B. International Society for Optics and
  Photonics, 2010.

\bibitem{poyade2009influence}
Matthieu Poyade, Arcadio Reyes-Lecuona, and Raquel Viciana-Abad.
\newblock Influence of binocular disparity in depth perception mechanisms in
  virtual environments.
\newblock In {\em New Trends on Human--Computer Interaction}, pages 13--22.
  Springer, 2009.

\bibitem{grinberg1994geometry}
Victor~S Grinberg, Gregg~W Podnar, and Mel Siegel.
\newblock Geometry of binocular imaging.
\newblock In {\em Stereoscopic Displays and Virtual Reality Systems}, volume
  2177, pages 56--65. International Society for Optics and Photonics, 1994.

\bibitem{akai2007depth}
Caitlin Akai.
\newblock {\em Depth perception in real and virtual environments: An
  exploration of individual differences}.
\newblock PhD thesis, School of Interactive Arts \& Technology-Simon Fraser
  University, 2007.

\bibitem{7012105}
Alexander Plopski, Yuta Itoh, Christian Nitschke, Kiyoshi Kiyokawa, Gudrun
  Klinker, and Haruo Takemura.
\newblock Corneal-imaging calibration for optical see-through head-mounted
  displays.
\newblock {\em IEEE Transactions on Visualization and Computer Graphics},
  21(4):481--490, 2015.

\bibitem{weier2018predicting}
Martin Weier, Thorsten Roth, Andr{\'e} Hinkenjann, and Philipp Slusallek.
\newblock Predicting the gaze depth in head-mounted displays using multiple
  feature regression.
\newblock In {\em Proceedings of the 2018 ACM Symposium on Eye Tracking
  Research \& Applications}, pages 1--9, 2018.

\bibitem{hirzle2019design}
Teresa Hirzle, Jan Gugenheimer, Florian Geiselhart, Andreas Bulling, and Enrico
  Rukzio.
\newblock A design space for gaze interaction on head-mounted displays.
\newblock In {\em Proceedings of the 2019 CHI Conference on Human Factors in
  Computing Systems}, pages 1--12, 2019.

\bibitem{elmadjian20183d}
Carlos Elmadjian, Pushkar Shukla, Antonio~Diaz Tula, and Carlos~H Morimoto.
\newblock 3d gaze estimation in the scene volume with a head-mounted eye
  tracker.
\newblock In {\em Proceedings of the Workshop on Communication by Gaze
  Interaction}, pages 1--9, 2018.

\bibitem{kocejko2009eye}
Tomasz Kocejko, Adam Bujnowski, and Jerzy Wtorek.
\newblock Eye-mouse for disabled.
\newblock In {\em Human-computer systems interaction}, pages 109--122.
  Springer, 2009.

\bibitem{gwon2014gaze}
Su~Yeong Gwon, Chul~Woo Cho, Hyeon~Chang Lee, Won~Oh Lee, and Kang~Ryoung Park.
\newblock Gaze tracking system for user wearing glasses.
\newblock {\em Sensors}, 14(2):2110--2134, 2014.

\bibitem{wibirama2017evaluating}
Sunu Wibirama, Hanung~A Nugroho, and Kazuhiko Hamamoto.
\newblock Evaluating 3d gaze tracking in virtual space: A computer graphics
  approach.
\newblock {\em Entertainment computing}, 21:11--17, 2017.

\bibitem{kocejko2015eye}
Tomasz Kocejko, Jacek Ruminski, Jerzy Wtorek, and Benoit Martin.
\newblock Eye tracking within near-to-eye display.
\newblock In {\em 2015 8th International Conference on Human System Interaction
  (HSI)}, pages 166--172. IEEE, 2015.

\bibitem{hosp2020remoteeye}
Benedikt Hosp, Shahram Eivazi, Maximilian Maurer, Wolfgang Fuhl, David Geisler,
  and Enkelejda Kasneci.
\newblock Remoteeye: An open-source high-speed remote eye tracker:
  Implementation insights of a pupil-and glint-detection algorithm for
  high-speed remote eye tracking.
\newblock {\em Behavior research methods}, 52(3), 2020.

\bibitem{li2018etracker}
Bin Li, Hong Fu, Desheng Wen, and WaiLun Lo.
\newblock Etracker: A mobile gaze-tracking system with near-eye display based
  on a combined gaze-tracking algorithm.
\newblock {\em Sensors}, 18(5):1626, 2018.

\bibitem{jung2016compensation}
Dongwook Jung, Jong~Man Lee, Su~Yeong Gwon, Weiyuan Pan, Hyeon~Chang Lee,
  Kang~Ryoung Park, and Hyun-Cheol Kim.
\newblock Compensation method of natural head movement for gaze tracking system
  using an ultrasonic sensor for distance measurement.
\newblock {\em Sensors}, 16(1):110, 2016.

\bibitem{fuchs2019smartlobby}
Stefan Fuchs, Nils Einecke, and Fabian Eisele.
\newblock Smartlobby: Using a 24/7 remote head-eye-tracking for content
  personalization.
\newblock In {\em Adjunct Proceedings of the 2019 ACM International Joint
  Conference on Pervasive and Ubiquitous Computing and Proceedings of the 2019
  ACM International Symposium on Wearable Computers}, pages 53--56, 2019.

\bibitem{lee20123d}
Ji~Woo Lee, Chul~Woo Cho, Kwang~Yong Shin, Eui~Chul Lee, and Kang~Ryoung Park.
\newblock 3d gaze tracking method using purkinje images on eye optical model
  and pupil.
\newblock {\em Optics and Lasers in Engineering}, 50(5):736--751, 2012.

\bibitem{hennessey2008noncontact}
Craig Hennessey and Peter Lawrence.
\newblock Noncontact binocular eye-gaze tracking for point-of-gaze estimation
  in three dimensions.
\newblock {\em IEEE transactions on biomedical engineering}, 56(3):790--799,
  2008.

\bibitem{abbott2012ultra}
William~Welby Abbott and Aldo~Ahmed Faisal.
\newblock Ultra-low-cost 3d gaze estimation: an intuitive high information
  throughput compliment to direct brain--machine interfaces.
\newblock {\em Journal of neural engineering}, 9(4):046016, 2012.

\bibitem{wibirama20143d}
Sunu Wibirama and Kazuhiko Hamamoto.
\newblock 3d gaze tracking on stereoscopic display using optimized geometric
  method.
\newblock {\em IEEJ Transactions on Electronics, Information and Systems},
  134(3):345--352, 2014.

\bibitem{chamberlain2007eye}
Laura Chamberlain.
\newblock Eye tracking methodology; theory and practice.
\newblock {\em Qualitative Market Research: An International Journal}, 2007.

\bibitem{daugherty2010measuring}
Brian~C Daugherty, Andrew~T Duchowski, Donald~H House, and Celambarasan
  Ramasamy.
\newblock Measuring vergence over stereoscopic video with a remote eye tracker.
\newblock In {\em Proceedings of the 2010 Symposium on Eye-Tracking Research \&
  Applications}, pages 97--100, 2010.

\bibitem{boev2012parameters}
Atanas Boev, Marianne Hanhela, Atanas Gotchev, Timo Utirainen, Satu
  Jumisko-Pyykk{\"o}, and Miska Hannuksela.
\newblock Parameters of the human 3d gaze while observing portable
  autostereoscopic display: a model and measurement results.
\newblock In {\em Multimedia on Mobile Devices 2012; and Multimedia Content
  Access: Algorithms and Systems VI}, volume 8304, page 830407. International
  Society for Optics and Photonics, 2012.

\bibitem{rozado2013mouse}
David Rozado.
\newblock Mouse and keyboard cursor warping to accelerate and reduce the effort
  of routine hci input tasks.
\newblock {\em IEEE Transactions on Human-Machine Systems}, 43(5):487--493,
  2013.

\bibitem{xia2007ir}
Dongshi Xia and Zongcai Ruan.
\newblock Ir image based eye gaze estimation.
\newblock In {\em Eighth ACIS International Conference on Software Engineering,
  Artificial Intelligence, Networking, and Parallel/Distributed Computing (SNPD
  2007)}, volume~1, pages 220--224. IEEE, 2007.

\bibitem{li2015gaze}
Jianfeng Li and Shigang Li.
\newblock Gaze estimation from color image based on the eye model with known
  head pose.
\newblock {\em IEEE Transactions on Human-Machine Systems}, 46(3):414--423,
  2015.

\bibitem{hennessey2006single}
Craig Hennessey, Borna Noureddin, and Peter Lawrence.
\newblock A single camera eye-gaze tracking system with free head motion.
\newblock In {\em Proceedings of the 2006 symposium on Eye tracking research \&
  applications}, pages 87--94, 2006.

\bibitem{pichitwong20163}
Wudthipong Pichitwong and Kosin Chamnongthai.
\newblock 3-d gaze estimation by stereo gaze direction.
\newblock In {\em 2016 13th International Conference on Electrical
  Engineering/Electronics, Computer, Telecommunications and Information
  Technology (ECTI-CON)}, pages 1--4. IEEE, 2016.

\bibitem{elmadjian20183d}
Carlos Elmadjian, Pushkar Shukla, Antonio~Diaz Tula, and Carlos~H Morimoto.
\newblock 3d gaze estimation in the scene volume with a head-mounted eye
  tracker.
\newblock In {\em Proceedings of the Workshop on Communication by Gaze
  Interaction}, pages 1--9, 2018.

\bibitem{lee2017estimating}
Youngho Lee, Choonsung Shin, Alexander Plopski, Yuta Itoh, Thammathip
  Piumsomboon, Arindam Dey, Gun Lee, Seungwon Kim, and Mark Billinghurst.
\newblock Estimating gaze depth using multi-layer perceptron.
\newblock In {\em 2017 International Symposium on Ubiquitous Virtual Reality
  (ISUVR)}, pages 26--29. IEEE, 2017.

\bibitem{essig2006neural}
Kai Essig, Marc Pomplun, and Helge Ritter.
\newblock A neural network for 3d gaze recording with binocular eye trackers.
\newblock {\em The International Journal of Parallel, Emergent and Distributed
  Systems}, 21(2):79--95, 2006.

\bibitem{hansen2009eye}
Dan~Witzner Hansen and Qiang Ji.
\newblock In the eye of the beholder: A survey of models for eyes and gaze.
\newblock {\em IEEE transactions on pattern analysis and machine intelligence},
  32(3):478--500, 2009.

\bibitem{sigut2010iris}
Jose Sigut and Sid-Ahmed Sidha.
\newblock Iris center corneal reflection method for gaze tracking using visible
  light.
\newblock {\em IEEE Transactions on Biomedical Engineering}, 58(2):411--419,
  2010.

\bibitem{sesma2012evaluation}
Laura Sesma, Arantxa Villanueva, and Rafael Cabeza.
\newblock Evaluation of pupil center-eye corner vector for gaze estimation
  using a web cam.
\newblock In {\em Proceedings of the symposium on eye tracking research and
  applications}, pages 217--220, 2012.

\bibitem{park2018learning}
Seonwook Park, Xucong Zhang, Andreas Bulling, and Otmar Hilliges.
\newblock Learning to find eye region landmarks for remote gaze estimation in
  unconstrained settings.
\newblock In {\em Proceedings of the 2018 ACM Symposium on Eye Tracking
  Research \& Applications}, pages 1--10, 2018.

\bibitem{duchowski2011measuring}
Andrew~T Duchowski, Brandon Pelfrey, Donald~H House, and Rui Wang.
\newblock Measuring gaze depth with an eye tracker during stereoscopic display.
\newblock In {\em Proceedings of the ACM SIGGRAPH symposium on applied
  perception in graphics and visualization}, pages 15--22, 2011.

\bibitem{wang2014online}
Rui~I Wang, Brandon Pelfrey, Andrew~T Duchowski, and Donald~H House.
\newblock Online 3d gaze localization on stereoscopic displays.
\newblock {\em ACM Transactions on Applied Perception (TAP)}, 11(1):1--21,
  2014.

\bibitem{mantiuk2011gaze}
Rados{\l}aw Mantiuk, Bartosz Bazyluk, and Anna Tomaszewska.
\newblock Gaze-dependent depth-of-field effect rendering in virtual
  environments.
\newblock In {\em International Conference on Serious Games Development and
  Applications}, pages 1--12. Springer, 2011.

\bibitem{padmanaban2017optimizing}
Nitish Padmanaban, Robert Konrad, Tal Stramer, Emily~A Cooper, and Gordon
  Wetzstein.
\newblock Optimizing virtual reality for all users through gaze-contingent and
  adaptive focus displays.
\newblock {\em Proceedings of the National Academy of Sciences},
  114(9):2183--2188, 2017.

\bibitem{pfeiffer2008evaluation}
Thies Pfeiffer, Marc~Erich Latoschik, and Ipke Wachsmuth.
\newblock Evaluation of binocular eye trackers and algorithms for 3d gaze
  interaction in virtual reality environments.
\newblock {\em JVRB-Journal of Virtual Reality and Broadcasting}, 5(16), 2008.


\bibitem{campbell1965optical}
FW~Campbell and DG~Green.
\newblock Optical and retinal factors affecting visual resolution.
\newblock {\em The Journal of physiology}, 181(3):576--593, 1965.

\end{thebibliography}
\end{document}